\documentclass[letterpaper]{article}

\usepackage{arxiv}
\usepackage{times}
\usepackage{helvet}
\usepackage{courier}
\usepackage[hyphens]{url}
\usepackage{graphicx}
\usepackage{natbib}
\usepackage{caption}
\usepackage{algorithm}
\usepackage{algorithmic}
\usepackage{color}
\usepackage{amsmath,amsfonts,bm}
\usepackage{booktabs}
\usepackage{newfloat}
\usepackage{listings}
\newcommand{\ours}{Singular Value Scaling}
\newcommand{\Fref}[1]{Figure~\ref{#1}}

\newcommand{\Tref}[1]{Table~\ref{#1}}

\usepackage{multirow} 
\usepackage{amsfonts}
\usepackage{booktabs}       
\usepackage{amsmath}
\usepackage{makecell}

\title{\ours: Efficient Generative Model Compression via Pruned Weights Refinement}
\author {
    Hyeonjin Kim and\
    Jaejun Yoo\thanks{Corresponding author.}
}
\affiliations {
    Laboratory of Advanced Imaging Technology (LAIT)\\
    Ulsan National Institute of Science and Technology (UNIST)\\
    \{hyeonjin.kim,\ jaejun.yoo\}@unist.ac.kr\\
}

\begin{document}

\maketitle

\begin{abstract}
While pruning methods effectively maintain model performance without extra training costs, they often focus solely on preserving crucial connections, overlooking the impact of pruned weights on subsequent fine-tuning or distillation, leading to inefficiencies. Moreover, most compression techniques for generative models have been developed primarily for GANs, tailored to specific architectures like StyleGAN, and research into compressing Diffusion models has just begun. Even more, these methods are often applicable only to GANs or Diffusion models, highlighting the need for approaches that work across both model types. In this paper, we introduce \ours~(SVS), a versatile technique for refining pruned weights, applicable to both model types. Our analysis reveals that pruned weights often exhibit dominant singular vectors, hindering fine-tuning efficiency and leading to suboptimal performance compared to random initialization. Our method enhances weight initialization by minimizing the disparities between singular values of pruned weights, thereby improving the fine-tuning process. This approach not only guides the compressed model toward superior solutions but also significantly speeds up fine-tuning. Extensive experiments on StyleGAN2, StyleGAN3 and DDPM demonstrate that SVS improves compression performance across model types without additional training costs. Our code is available at: \url{https://github.com/LAIT-CVLab/Singular-Value-Scaling}.
\end{abstract}

\section{Introduction}

Generative models like Generative Adversarial Networks (GANs) \cite{goodfellow2014generative} and Diffusion models \cite{ho2020denoising} have achieved remarkable performance across various computer vision tasks such as image generation \cite{Karras2019stylegan2, ho2020denoising, Wang_2022_CVPR}, image editing \cite{CLIPInverter, Pehlivan_2023_CVPR, kawar2023imagic, zhang2023sine}, even video generation \cite{stylegan-v, ho2022video, blattmann2023videoldm} and 3D generation \cite{Chan2022, Karnewar_2023_CVPR}. The impressive performance of these generative models, however, often comes at the cost of high memory and computational demands, limiting their real-world applicability. To address these issues, several model compression technique for generative models have been proposed \cite{liu2021content, xu2022mind, chung2024diversity, fang2023structural}.

Model compression typically involves two steps: 1) pruning to reduce model size while retaining essential pre-trained knowledge, and 2) fine-tuning to restore performance. Among these, effective pruning has received significant attention for its ability to enhance following fine-tuning step. The retained pre-trained knowledge from pruning can enhance fine-tuning, leading to improved performance and faster convergence without additional training costs. 
While previous pruning methods \cite{chung2024diversity, fang2023structural} effectively maintain model performance, they often focus solely on preserving crucial connections in the pre-trained model, overlooking the impact of pruned weights on subsequent process, leading to inefficient fine-tuning or distillation. This issue becomes more severe as the model's capacity decreases, and in some cases, the pruned weights results in worse performance compared to random initialization; i.e. slow convergence speed and lower performance. Therefore, addressing these factors is essential for achieving more efficient generative models.

Our key observation is that pruned weights often exhibit dominant singular vectors, which results in a large disparity between the largest and smallest singular values. The presence of such dominant singular vectors significantly impacts the model's forward propagation, overshadowing the contributions of minor singular vectors, and the overall fine-tuning process is dominated by these few dominant singular vectors. We find that this could limit the exploration of diverse weight space. Therefore, ensuring a balanced contribution among the singular vectors within pruned weights at initialization may offer a potential solution to the inefficiencies observed in the fine-tuning process.

In this paper, we introduce a simple yet effective refinement technique called \ours~(SVS) to enhance the efficiency of fine-tuning pruned weights. The dominant singular values is scaled down to reduce the disparity between the largest singular values and smallest singular values while preserving the relative order of the singular values at initialization. In this way, the refined pruned weights makes the fine-tuning process easier compared to directly using the pruned weights by ensuring all singular vectors contribute more evenly at the beginning of fine-tuning. Note that since our method focuses on the knowledge within each weight and is independent of specific architectures, it can be applied to both GANs and diffusion models. 

We conduct extensive experiments with representative generative model architectures across various datasets, including StyleGAN2, StyleGAN3, and Denoising Diffusion Probabilistic Model (DDPM) on CIFAR10, CelebA-HQ, FFHQ, and LSUN-Church. The results demonstrate that our method enhances the fine-tuning process, leading to both faster convergence and improved solutions for the compressed model without additional training cost. Our contributions can be summarized as follows:
\begin{itemize}
    \item We propose a simple yet effective method that enhances the efficacy of fine-tuning process of pruned generative models without additional training cost. 
    \item By simply scaling down the singular values of pruned weights at initialization, our method, \ours~(SVS) enables the compressed model to converge faster and achieve superior performance than using the existing baseline methods. 
    \item We are the first to provide a general method for generative model compression that can be applied to both model types of GANs and Diffusion models. 
\end{itemize}

\section{Related Works}
\subsection{StyleGAN Compression}

Recently, several studies \cite{liu2021content, xu2022mind, chung2024diversity} have been proposed for compressing unconditional GANs, particularly the StyleGAN family \cite{Karras2019stylegan2, karras2021alias}, which are the state-of-the-art models in the area. CAGAN \cite{liu2021content} introduced the first framework for pruning pre-trained StyleGAN models and fine-tuning them via pixel-level and feature-level knowledge distillation. Following this, StyleKD \cite{xu2022mind} further enhanced the fine-tuning process by incorporating a specialized relation loss tailored for StyleGAN. In particular, CAGAN introduced a content-aware pruning technique that preserves connections in the pre-trained model crucial for the semantic part of generated images, while StyleKD tackled output discrepancies problem by inheriting only the pre-trained model's mapping network and randomly initializing the synthesis network with smaller size. More recently, DCP-GAN \cite{chung2024diversity} proposed a diversity-aware channel pruning technique, which preserves connections in the synthesis network that contribute to sample diversity, significantly enhancing both the diversity of generated images and the training speed.

\subsection{Diffusion Model Compression}

Diffusion models have garnered significant attention for their stable training and impressive generative capabilities. However, their performance incurs high computational costs due to iterative sampling. 
To enhance efficiency, various sampling techniques have been developed to reduce the number of required iterations \cite{song2021denoising, lu2022dpmsolver, zheng2023fast}. Orthogonal to the these sampling techniques, Diff-Prune \cite{fang2023structural} introduced a seminal work for reducing computational costs by compressing Diffusion models. This demonstrated that by primarily pruning connections involved in less important diffusion steps, a reduced-size Diffusion model with minimal performance loss can be achieved in significantly fewer training iterations compared to the original-sized model.

\subsection{Weight Initialization and Trainability}

Weight initialization is essential for efficient training in deep learning. Glorot \cite{glorot2010understanding} and He initialization \cite{he2015delving} are widely used to maintain activation variance and ensure stable optimization, particularly in feed-forward networks with ReLU activations. Orthogonal initialization \cite{saxe2014exact}, which ensures that all singular values of weights are 1, achieves dynamical isometry, the ideal state for trainability. In classifier pruning, TPP \cite{wang2023trainability} recently highlighted the importance of preserving trainability in pruned networks to support effective performance recovery.

\section{Analysis}

\begin{figure*}[t]
\centering
\includegraphics[width=\linewidth]{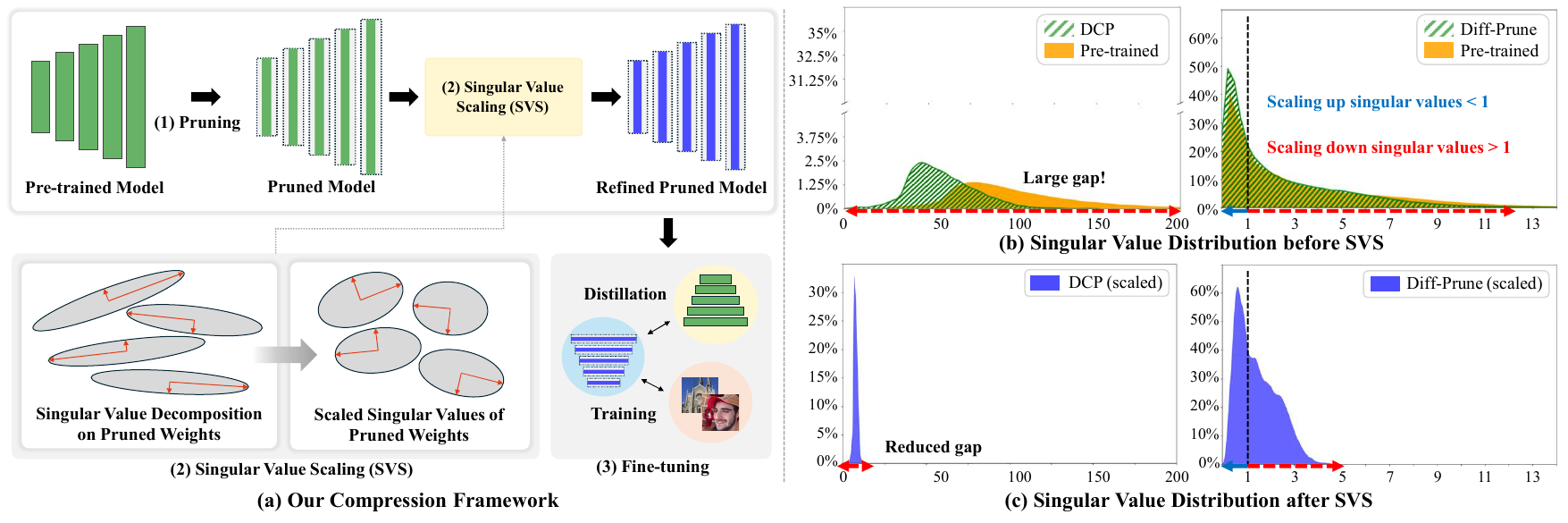}
\caption{(a) A schematic overview of our compression framework. Unlike existing model compression scheme, we additionally perform pruned weights refining step for efficient model fine-tuning. First, we prune the pre-trained model. Next, we compute singular values of the weights of the pruned model. Then, we refine the pruned weights by scaling down the singular values with large magnitudes and scaling up the singular values with small magnitudes (\textit{Singular Value Scaling, SVS}). Finally, the refined pruned model are fine-tuned. (b), (c) The singular value distribution before and after singular value scaling in the pruned weights by the method of DCP-GAN and Diff-Prune. 
The x-axis represents singular values and the y-axis represents the density. When applied SVS, the singular value disparity reduces, balancing the power of each singular vector in the pruned weights.}
\label{figure:framework}
\end{figure*}

\begin{figure}[t]
\centering
\includegraphics[width=\columnwidth]{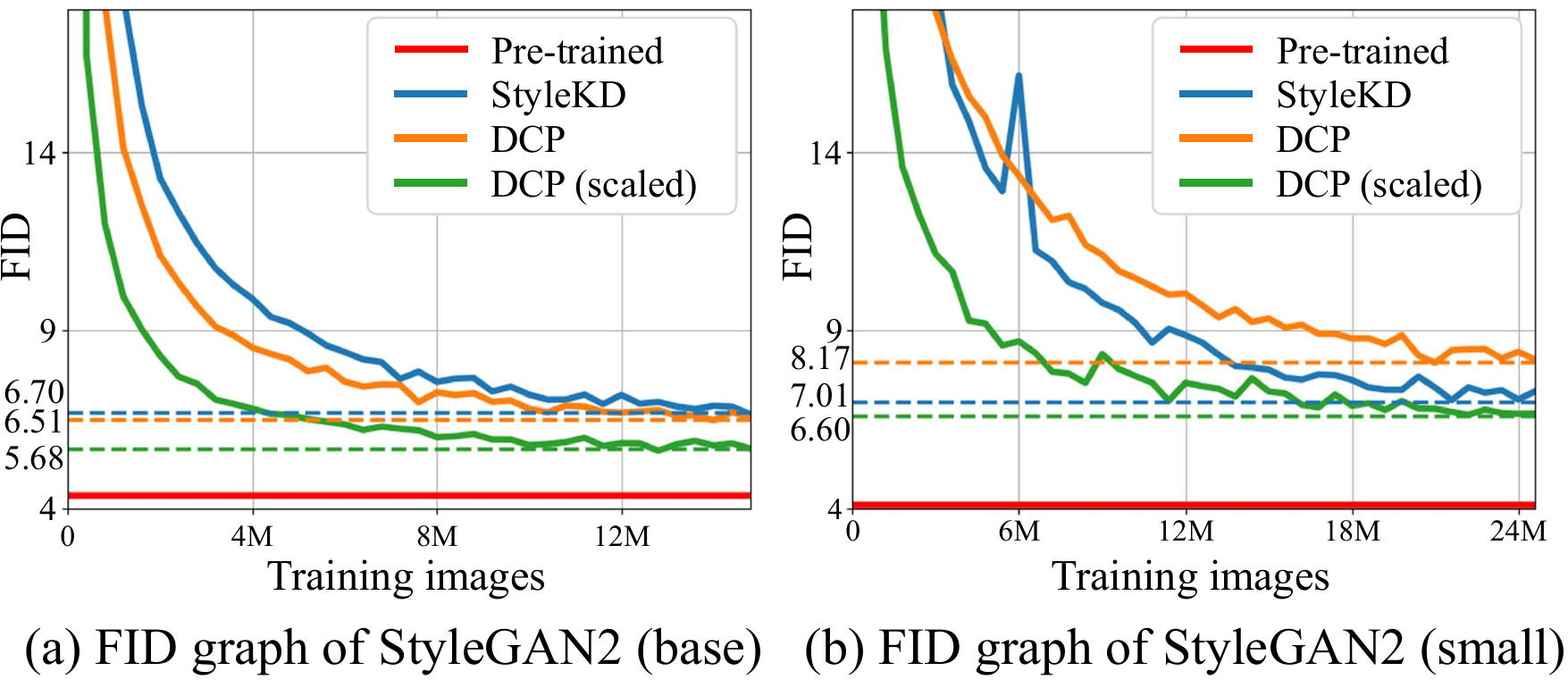}
\caption{FID convergence graph in different StyleGAN2 Architectures compressed by different methods. The x-axis represents the number of images shown to the discriminator. Solid line represents FID with respect to the training images, while dashed line represents the best FID of the compressed model by the corresponding method. ``scaled'' means that our \textit{\ours} is applied to the pruned weights.} 
\label{figure:stylegan-fid-convergence}
\end{figure}

\subsection{suboptimal Results with Pruned Weights}
Previous works in the field of generative model compression \cite{chung2024diversity, fang2023structural} have shown that well-designed pruning techniques can lead compressed models to better solutions compared to their randomly initialized counterparts, while reaching the same performance more quickly ($\times2.5$ for StyleGAN2, $\times8.0$ for DDPM). This is because pruned weights provide the compressed model with an initialization state that preserves the pre-trained model's performance, starting with better performance compared to the random counterpart. This initial gain continues throughout the fine-tuning process, ultimately resulting in a better-performing compressed model. However, we observe that as fine-tuning progresses, models initialized with pruned weights converge more slowly compared to those initialized randomly (see \Fref{figure:stylegan-fid-convergence}: orange line (DCP-GAN) vs. blue line (StyleKD)). This issue worsens as the model size decreases, resulting in suboptimal solutions, which indicates that the pruned weights themselves contain factors that hinder fine-tuning efficiency. 

\subsection{Analyzing Pruned Weights with SVD}

We analyze the learned prior of pruned weights inherited from the pre-trained model by employing Singular Value Decomposition (SVD), which is widely adopted for low-rank approximation by identifying important basis from the weights \cite{denton2014exploiting, zhang2015efficient, girshick2015fast, yoo2019photorealistic}. For a weight matrix $W\in \mathcal{R}^{m\times n}\ (m<n)$, we can decompose it using SVD:
$$
W=U\Sigma V^T=\sum^m_{i=1}\sigma_i\vec u_i\vec v_i^T
$$
where $U=[\vec u_1, \vec u_2,\cdots,\vec u_m]\in\mathcal{R}^{m\times m}$ and $V=[\vec v_1, \vec v_2, \cdots, \vec v_n]\in\mathcal{R}^{n\times n}$ are orthogonal matrices and $\Sigma=diag(\sigma_1,\sigma_2,\cdots, \sigma_m)\in\mathcal{R}^{m\times n}$ is a diagonal matrix with singular values on the diagonal. Here, for a fully connected layer, $W\in\mathcal{R}^{c_{out}\times c_{in}}$. For a convolutional layer, $W\in\mathcal{R}^{c_{out}\times(c_{in}\times k\times k)}$, where $k$ is the kernel size. For a convolutional layer, we consider the weight is flattened. Each singular value represents the influence of its corresponding singular vector within the weight. 
The most notable observation is the large gap between the largest and smallest singular values of pruned weights ($\sim \times 100$). This implies that the forward and backward propagations of the weights are heavily influenced by these dominant singular vectors. This can potentially bias the compressed model towards these singular vectors during training, severely limiting diverse exploration in the weight space \cite{saxe2014exact, wang2023trainability}. 

\section{Method}

\subsection{Scaling Singular Values of Pruned Weights}
Based on our observation, we propose ``\ours~(SVS)", to refine pruned weights to enhance fine-tuning efficiency. Our primary goal is to reduce the gap between the singular values of pruned weights. In the pruned weights, dominant singular vectors tend to have significantly larger singular values compared to smaller ones, and this gap increases as the values grow. Since all bases in pruned weights contain important knowledge from the pre-trained model, we prevent any single basis from dominating to let these bases contribute equally at the beginning of training. To achieve this, we simply scale the singular values using the ``square root function",
$$
W_{scaled}=U\Sigma_{scaled} V^T=\sum^m_{i=1}\sqrt{\sigma_i}\vec u_i\vec v_i^T
$$
Here, $U$ and $V$ remain unchanged, and $\Sigma_{scaled}=diag(\sqrt{\sigma_1},\sqrt{\sigma_2},\cdots,\sqrt{\sigma_m})\in\mathcal{R}^{m\times n}$. Square root function has several good properties: 1) Since singular values are non-negative, it maps non-negative values to non-negative values, 2) For $\sigma_i<1$, it increases the value, while for $\sigma_i>1$, it decreases the value more as the value grows. 3) In the positive domain, it is monotonically increasing, thereby maintaining the relative order of singular values. By scaling the singular values of pruned weights in this manner, we can preserve the original bases while balancing their relative contributions. This balanced contribution helps the compressed model fully leverage the pre-trained model's knowledge, facilitating a more effective path to the optimal solution. 
Alternative functions, such as $\log(x+1)$ or $\vert\log(x)\vert$, could also achieve similar effects. We provide ablation study on these functions in the Experiments section.

\subsection{Scaling Bias with respect to the Scaled Weights}

Since biases and weights are learned simultaneously, we need to consider the impact of scaling the singular values of the weights on the corresponding biases. Let us consider the following linear equation,
$
y=Wx+b, 
$
where $W\in\mathcal R^{m\times n}$ is the weights, $x\in\mathcal R^{n}$ is the input and $b\in\mathcal R^{m}$ is the bias. We can factor the equation with respect to $W$ as follows:
$$
y=W(x+W^\dagger b)=W(x+V\Sigma^{-1}U^Tb)
$$
where $W^\dagger\in\mathcal R^{n\times m}$ is the Moore-Penrose inverse \cite{petersen2012linear} and $\Sigma^{-1}=diag(\sigma^+_1, \sigma^+_2,\cdots,\sigma^+_m)\in\mathcal R^{n\times m}$ with
$$
\sigma^+_{i} = 
\begin{cases} 
\frac{1}{\sigma_i}, & \text{if } \sigma_i \neq 0 \\
0, & \text{if } \sigma_i = 0
\end{cases}
$$

\noindent By representing the bias $b$ as $b=|b|\vec\beta$, where $|\cdot|$ is the vector norm and $\vec\beta$ is the normalized vector of $b$, it becomes:
$$
y=W(x+V\Sigma^{-1}U^Tb)=W(x+V(|b|\Sigma^{-1})U^T\vec\beta)
$$
where $|b|\Sigma^{-1}=diag(|b|\sigma^+_1, |b|\sigma^+_2,\cdots,|b|\sigma^+_m)$. Thus, for $\sigma_i>0$, when scaling the biases, instead of only scaling $\sigma_i$, we must also scale $b$:
$
b_{scaled}=b/\sqrt{|b|},
$
which leads to:
$$
|b_{scaled}|/\sigma_{i,scaled}=\sqrt{|b|/\sigma_i}.
$$

\section{Experiments}

\begin{table*}[th!]
\centering\resizebox{\linewidth}{!}
{
    \begin{tabular}{c|c|c|cc|ccccc}
    \hline
    Dataset & Method & Arch & Params  $\downarrow$ & FLOPs $\downarrow$ & FID $\downarrow$ & P $\uparrow$ & R $\uparrow$ & D $\uparrow$ & C $\uparrow$ \\ \hline \hline
    
    \multirow{12}{*}{FFHQ}
    
    & Teacher & \multirow{4}{*}{\text{base}} & 30.0M & 45.1B & 4.29 & 0.762 & 0.561 & 1.061 & 0.845 \\ 
    & StyleKD \cite{xu2022mind} &  & \multirow{3}{*}{5.6M} & \multirow{3}{*}{4.1B} & 6.70 & 0.711 & \underline{0.551} & 0.879 & \underline{0.786} \\
    & DCP-GAN \cite{chung2024diversity} &  &  &  & \underline{6.51} & \underline{0.712} & \textbf{0.556} & \underline{0.884} & 0.785 \\ 
    & \textbf{DCP-GAN (scaled)} &  &  &  & \textbf{5.68} & \textbf{0.740} & 0.530 & \textbf{0.981} & \textbf{0.806} \\  \cline{2-10}
    
    & Teacher & \multirow{6}{*}{\text{small}} & 24.7M & 14.9B & 4.02 & 0.769 & 0.555 & 1.095 & 0.854 \\ 
    & GS \cite{wang2020ganslimming}&  & \multirow{6}{*}{4.9M} & \multirow{6}{*}{1.3B} & 10.23 & 0.702 & 0.430 & 0.845 & 0.721 \\
    & CAGAN \cite{liu2021content}&   &  & & 9.23 & 0.685 & 0.500 & 0.760 & 0.722 \\
    & StyleKD \cite{xu2022mind} &   &  &  & \underline{7.01} & \underline{0.707} & \textbf{0.543} & \textbf{0.900} & \textbf{0.783} \\
    & DCP-GAN \cite{chung2024diversity} &  &  &  & 8.17 & 0.680 & \underline{0.539} & 0.787 & 0.741 \\ 
    & \textbf{DCP-GAN (scaled)} &  & &  & \textbf{6.60} & \textbf{0.719} & 0.532 & \underline{0.874} & \underline{0.778} \\ \hline
    
    \multirow{8}{*}{Church}
    
    & Teacher & \multirow{4}{*}{\text{base}} & 30.0M & 45.1B & 3.97 & 0.698 & 0.553 & 0.849 & 0.823  \\ 
    & StyleKD \cite{xu2022mind} &  &\multirow{3}{*}{5.6M} & \multirow{3}{*}{4.1B} & 5.72 & \underline{0.703} & 0.460 & \underline{0.845} & 0.777 \\
    & DCP-GAN \cite{chung2024diversity} &  & &  & \underline{4.80} & 0.692 & \textbf{0.505} & 0.810 & \underline{0.790} \\ 
    & \textbf{DCP-GAN (scaled)} &  &  &  & \textbf{4.62} & \textbf{0.716} & \underline{0.501} & \textbf{0.878} & \textbf{0.812} \\  \cline{2-10}
    & Teacher & \multirow{4}{*}{\text{small}} & 24.7M & 14.9B & 4.37 & 0.696 & 0.512 & 0.815 & 0.809 \\ 
    & StyleKD \cite{xu2022mind} &  & \multirow{3}{*}{1.38B} & \multirow{3}{*}{1.38B} & \underline{5.99} & \underline{0.698} & \underline{0.446} & \underline{0.848} & \underline{0.777} \\
    & DCP-GAN \cite{chung2024diversity} &  & &  & 6.30 & 0.687 & 0.435 & 0.793 & 0.762 \\ 
    & \textbf{DCP-GAN (scaled)} &  &  &  & \textbf{5.18} & \textbf{0.719} & \textbf{0.464} & \textbf{0.934} & \textbf{0.805} \\  \hline

    \end{tabular}
}

\caption{Quantitative Results on StyleGAN2 compression. We report training results of previous methods on FFHQ-256 and LSUN Church-256 datasets. ``Params'' and ``FLOPs'' refer to the number of parameters in generator and floating-point operations respectively. ``P'' and ``R'' denote precision and recall metrics, while ``D'' and ``C'' represent density and coverage metrics. ``base'' and ``small'' denote the StyleGAN2 architecture with different capacity. ``scaled'' means that our \textit{\ours} is applied to the pruned weights. All reported metrics are trained results with the official StyleGAN2 implementation.}
\label{table:fid-stylegan2}
\end{table*}

\begin{table}[ht!]
\centering\resizebox{\linewidth}{!}
{
    \begin{tabular}{c|ccccc}
    \hline
    Method & FID $\downarrow$ & P $\uparrow$ & R $\uparrow$ & D $\uparrow$ & C $\uparrow$ \\ 
    \hline \hline
    Teacher & 4.76 & 0.736 & 0.593 & 0.955 & 0.817 \\ 
    DCP-GAN & 9.88 & 0.694 & 0.484 & 0.780 & 0.720 \\
    \makecell{\textbf{DCP-GAN} \\ \textbf{(scaled)}} & \textbf{8.38} & \textbf{0.702} & \textbf{0.531} & \textbf{0.808} & \textbf{0.748} \\  
    \hline
    \end{tabular}
}
\caption{Quantitative Results on StyleGAN3 Compression.}
\label{table:fid-stylegan3}
\end{table}

\begin{table*}[ht!]
{\centering\resizebox{\linewidth}{!}
{

    \begin{tabular}{ccccccccccc}
    \hline
    \multicolumn{11}{c}{\textbf{DDPM CIFAR-10 32$\times$32 (100 DDIM steps)}} \\ \hline
    Method & Channel Sparsity $\uparrow$ & Params $\downarrow$ & MACs $\downarrow$ & FID $\downarrow$ & SSIM $\uparrow$ & P $\uparrow$ & R $\uparrow$ & D $\uparrow$ & C $\uparrow$ &  Train Steps $\downarrow$ \\ \hline 
    pre-trained & 0\% & 35.7M & 6.1G & 4.19 & 1.000 & 0.759 & 0.672 & 0.994 & 0.905 & 800K \\ \hline
    Diff-Prune$^\dagger$ & \multirow{3}{*}{30\%} & \multirow{3}{*}{19.8M} & \multirow{3}{*}{3.4G} & 5.29 & \textbf{0.932} & N/A & N/A & N/A & N/A & 100K \\ 
    Diff-Prune & & & & 5.49 & 0.931 & \textbf{0.742} & \textbf{0.671} & 0.935 & \textbf{0.875} & \multirow{2}{*}{160K} \\
    \textbf{Diff-Prune (scaled)} & & & & \textbf{5.14} & 0.923 & 0.740 & 0.668 & \textbf{0.936} & 0.872 & \\ \hline
    Diff-Prune & \multirow{2}{*}{50\%} &  \multirow{2}{*}{8.96M} & \multirow{2}{*}{1.5G} & 7.77 & \textbf{0.914} & \textbf{0.720} & 0.670 & \textbf{0.840} & 0.815 & \multirow{2}{*}{360K} \\
    \textbf{Diff-Prune (scaled)} & &  &  & \textbf{7.41} & 0.912 & 0.718 & \textbf{0.671} & 0.836 & \textbf{0.816} & \\ \hline
    Diff-Prune & \multirow{2}{*}{70\%} & \multirow{2}{*}{5.12M} & \multirow{2}{*}{1.0G} & 9.87 & 0.906 & \textbf{0.702} & 0.665 & \textbf{0.777} & 0.767 & \multirow{2}{*}{600K}\\
    \textbf{Diff-Prune (scaled)} &  & & & \textbf{9.38} & \textbf{0.907} & 0.699 & \textbf{0.670} & 0.774 & \textbf{0.772} & \\ \hline \hline
    \end{tabular}
    
}
\resizebox{\linewidth}{!}
{

    \begin{tabular}{ccccccccccc}
    \multicolumn{11}{c}{\textbf{DDPM CelebA-HQ 64 × 64 (100 DDIM steps)}} \\ \hline
    Method & Channel Sparsity $\uparrow$ & Params $\downarrow$ & MACs $\downarrow$ & FID $\downarrow$ & SSIM $\uparrow$& P $\uparrow$ & R $\uparrow$ & D $\uparrow$ & C $\uparrow$ & Train Steps $\downarrow$ \\ \hline 
    pre-trained & 0\% & 78.7M & 23.9G & 6.48 & 1.000 & 0.812 & 0.587 & 1.069 & 0.884 & 500K \\ \hline
    Diff-Prune$^{\dagger}$ &  \multirow{3}{*}{30\%} &  \multirow{3}{*}{43.7M} & \multirow{3}{*}{13.3G} & 6.24 & 0.885 & N/A & N/A & N/A & N/A & 100K \\ 
    Diff-Prune & & & & 6.17 & \textbf{0.949} & \textbf{0.803} & 0.591 & \textbf{1.043} & 0.881 & \multirow{2}{*}{100K} \\
    \textbf{Diff-Prune (scaled)} & &  & & \textbf{5.57} & 0.938 & 0.793 & \textbf{0.619} & 1.011 & \textbf{0.889} &  \\ \hline
    Diff-Prune &  \multirow{2}{*}{50\%} & \multirow{2}{*}{19.7M} & \multirow{2}{*}{6.0G} & 5.32 & \textbf{0.926} & \textbf{0.773} & 0.618 & \textbf{0.928} & 0.875 & \multirow{2}{*}{200K} \\
    \textbf{Diff-Prune (scaled)} &  & &  & \textbf{4.66} & 0.921 & 0.766 & \textbf{0.634} & 0.925 & \textbf{0.885} & \\ \hline
    Diff-Prune &  \multirow{2}{*}{70\%} & \multirow{2}{*}{9.25M} & \multirow{2}{*}{4.2G} & 6.30 & \textbf{0.913} & \textbf{0.772} & 0.584 & \textbf{0.928} & 0.857 & \multirow{2}{*}{240K} \\
    \textbf{Diff-Prune (scaled)} &  & &  & \textbf{5.04} & 0.905 & 0.752 & \textbf{0.625} & 0.873 & \textbf{0.864} & \\ \hline \hline
    \end{tabular}
    
}
}
\caption{Quantitative results on DDPM compression. ``$\dagger$'' represents reported results in the paper. ``N/A'' indicates the inability to evaluate due to the inaccessibility of the weights used in the paper. ``Channel Sparsity'' refers the ratio of removed channels from the pre-trained model and ``Params'' and ``MACs'' are resulting number of parameters and multiply-accumulate of the compressed model. ``SSIM'' estimates the similarity between generated sample of pre-trained and compressed models given same noise. ``P'' and ``R'' denote precision and recall metrics, while ``D'' and ``C'' represent density and coverage metrics. ``scaled'' means that our \textit{\ours} is applied to the pruned weights.}

\label{table:diffusion-compression}
\end{table*}

\subsection{Experimental Setup}
\paragraph{Baselines.}
To evaluate our method, we conducted experiments on StyleGAN2, which is the most extensively studied model in the field of generative model compression. Additionally, to validate our method on generative models with different architectures, we conducted experiments on StyleGAN3 and the Denoising Diffusion Probabilistic Model (DDPM), which, along with StyleGAN2, are representative models in the field of generative modeling.
\paragraph{Implementation details.}
For the StyleGAN2 compression, we re-implement previous StyleGAN2 compression methods, which were implemented on the unofficial StyleGAN2 implementation\footnote{https://github.com/rosinality/stylegan2-pytorch}, on the official StyleGAN2 implementation\footnote{https://github.com/NVlabs/stylegan2-ada-pytorch} because the official implementation provides more optimized StyleGAN2 training settings and are easy to use. 
we apply our method to the weights pruned by DCP-GAN \cite{chung2024diversity}, which is the state-of-the-art pruning method in StyleGAN2 compression. We mainly compare our method against StyleKD \cite{xu2022mind} and DCP-GAN because they are the methods that demonstrate the best performance in StyleGAN2 compression. We use two StyleGAN2 architectures: the optimized architecture provided by the official implementation (denoted as ``StyleGAN2 (small)") and the original structure used in previous works (denoted as ``StyleGAN2 (base)"). Experiments are conducted on the FFHQ \cite{stylegan1} and LSUN Church \cite{yu2015lsun} datasets. Following DCP-GAN, we use a facial mask as the content mask for the FFHQ dataset and a uniform mask (all pixel values are assigned values 1) for LSUN Church dataset. 
For StyleGAN3 compression, we implement our method based on the DCP-GAN implementation\footnote{https://github.com/jiwoogit/DCP-GAN/tree/stylegan3}. Similar to StyleGAN2 compression, we apply our method to the pruned weights using the DCP-GAN method for our experiments. We follow the experimental setup of DCP-GAN and train StyleGAN3 on the FFHQ dataset until the discriminator see 10 million images for both DCP-GAN and our method. 
For DDPM compression, we build upon Diff-Prune \cite{fang2023structural}\footnote{https://github.com/VainF/Diff-Pruning}, a foundational work in Diffusion model pruning. We train pruned DDPMs using the CIFAR10 \cite{alex2009learning} and CelebA-HQ \cite{liu2015faceattributes} datasets, following the experimental setup of Diff-Prune. Unlike Diff-Prune, which tested only on a 30\% channel sparsity, we explore more extreme levels of sparsity in our experiments, including 50\% and 70\% channel sparsity. We provide more detailed implementation in the supplementary material.

\paragraph{Evaluation metrics.}
For StyleGAN compression, we evaluate our method using the Fr{\'e}chet Inception Distance (FID) \cite{heusel2017gans} and Precision and Recall (P\&R) \cite{kynkaanniemi2019improved}. We also measure Density and Coverage (D\&C) \cite{naeem2020reliable}, which are more robust to outliers than Precision and Recall. Here, Precision and Density evaluate the fidelity of generated samples, while Recall and Coverage assess their diversity. To calculate FID, we use all real samples from each dataset and 50K fake samples. For P\&R and D\&C calculation, we use 50K fake samples and all real samples from FFHQ, and 50K real samples from LSUN Church due to computational costs. For DDPM compression, in addition to FID, we employ Structural Similarity (SSIM) \cite{wang2004image} to evaluate consistency with the pre-trained model, following the Diff-Prune. The SSIM score is measured between images generated by the pre-trained model and the compressed model, given the identical noise inputs. We use a 100-step DDIM sampler \cite{song2021denoising} for sampling.

\subsection{Analysis on Convergence Speed}

\Fref{figure:stylegan-fid-convergence} visualizes the FID convergence graph for two different StyleGAN2 architectures (StyleGAN2 (base) and StyleGAN2 (small)) in the FFHQ dataset. As shown in \Fref{figure:stylegan-fid-convergence} (a) (StyleGAN2 (base)), although DCP-GAN converges to better performance than StyleKD, the FID convergence speed of DCP-GAN noticeably slows down in the later stages of training. This issue becomes even more severe as the model's capacity decreases. In \Fref{figure:stylegan-fid-convergence} (b), this slowdown allows StyleKD to surpass DCP-GAN early in fine-tuning, leading DCP-GAN to a suboptimal point. These results indicate that fine-tuning a model naively initialized with the pruned weights can lead to significantly poor outcomes, depending on the model's capacity. In contrast, when pruned weights are refined using SVS, the model is much easier to fine-tune. It not only fits faster across most intervals but also converges to a significantly better solution compared to previous methods. 
\subsection{Quantitative Results on StyleGAN2 compression}

\Tref{table:fid-stylegan2} shows results on StyleGAN2 compression on different StyleGAN2 architectures in FFHQ and LSUN Church datasets. For FID, our method outperforms previous compression methods with a clear margin in all experiments. It generally outperforms previous methods in Precision and Density metrics, and matches or exceeds their performance in Recall and Coverage metrics. These results show that our method enables the compressed model to better match the distribution compared to previous methods, leading to higher quality and more diverse samples. 
One notable observation is that when the capacity of StyleGAN2 is sufficient (base), DCP-GAN consistently outperforms StyleKD, and as the capacity decreases (small), DCP-GAN's performance lags behind that of StyleKD, indicating that the pruned weights from DCP-GAN are harder to fine-tune. In contrast, our method provides a more effective and refined initialization, resulting in improved final performance.

\subsection{Quantitative Results on StyleGAN3 compression}

\Tref{table:fid-stylegan3} presents the results of StyleGAN3 compression. The model initialized with our refined pruned weights outperforms the DCP-GAN across all metrics. This result shows that, similar to StyleGAN2, refining pruned weights facilitates easier fine-tuning of the model, enabling it to generate more plausible and diverse images. 

\subsection{Quantitative Results on DDPM compression}

\Tref{table:diffusion-compression} presents the results of DDPM compression.\footnote{During the rebuttal phase, we found a mismeasurement in Precision, Recall, Density, and Coverage in \Tref{table:diffusion-compression}. After correcting it, the results are now more consistent, reaffirming that \ours~preserves diversity well.}   1q` For DDPM compression, we compare our method with the best-performing baseline during fine-tuning. Diff-Prune achieves higher SSIM scores, indicating closer alignment with the pre-trained model. This is expected, as our method refines pruned weights to balance knowledge preservation with easier fine-tuning. Consequently, our method consistently achieves better FID scores across all pruning ratios, indicating more effective distribution matching. Additionally, the modest drop in SSIM indicates that our method still effectively retains contextual knowledge, even as it further relaxes the knowledge of the pre-trained model.

\begin{figure*}[t]
\centering
\includegraphics[width=\linewidth]{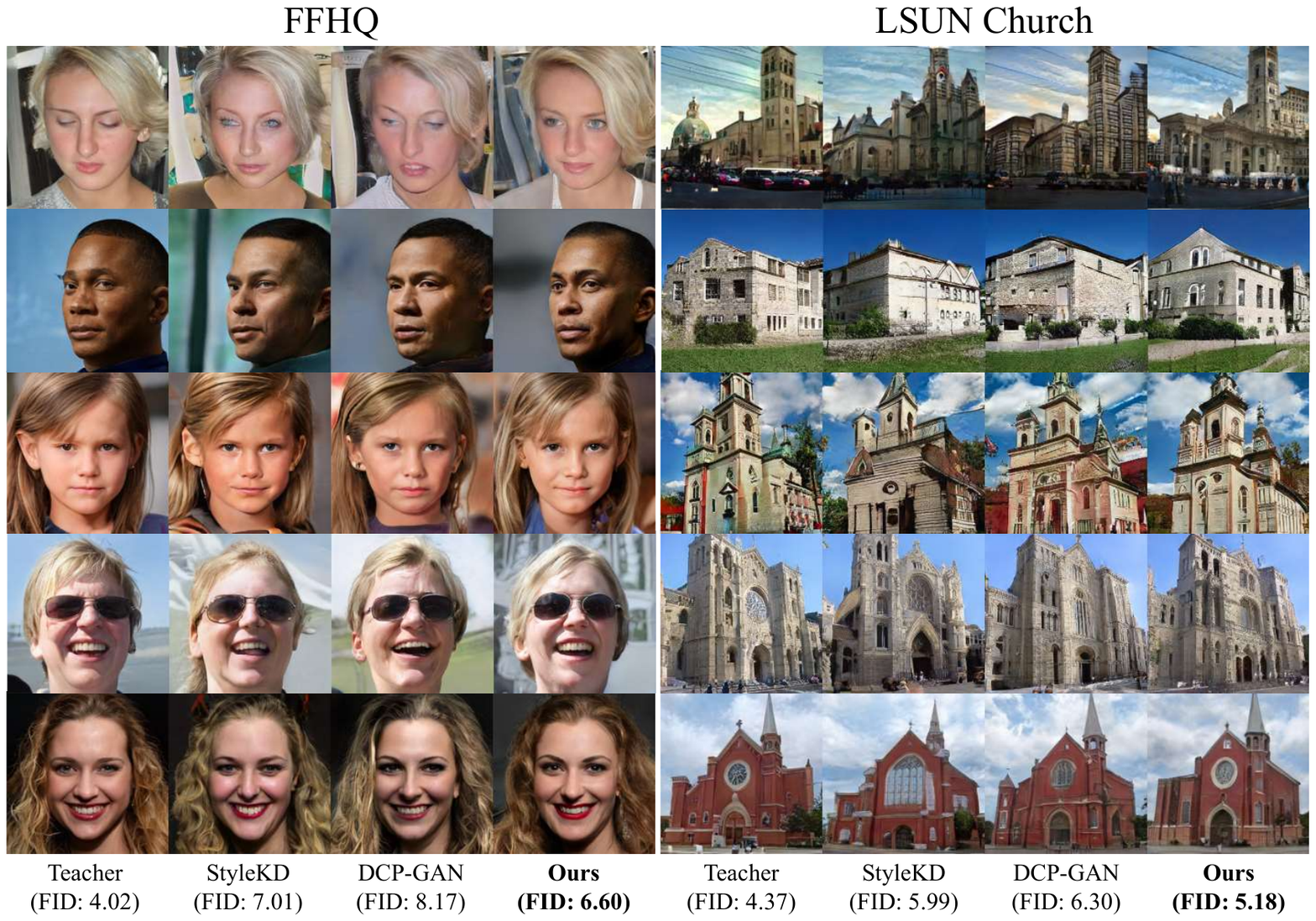}
\caption{Qualitative results on FFHQ and LSUN Church datasets. Samples in each row are generated from same noise vector with StyleGAN2 (small), which is compressed using different compression methods with channel sparsity 70\%. ``Ours'' denotes the compressed model with DCP-GAN refined by \textit{\ours}.}
\label{figure:qualitative-results}
\end{figure*}

\subsection{Qualitative Results on StyleGAN2 compression}

\Fref{figure:qualitative-results} shows samples generated by StyleGAN2 (small) compressed using different methods from the same noise vector. Despite all compressed models being trained with the same loss, the characteristics of the generated samples vary depending on the weight initialization and frequently exhibit artifacts. For example, the model compressed by StyleKD often misses contextual details, such as face shape (second and fourth row). DCP-GAN, which prunes the pre-trained model to maximize diversity, tends to focus heavily on details like wrinkles and hair, while often failing to restore the overall appearance of the face. Additionally, as seen in the third column, first row of LSUN Church, DCP-GAN tends to generate excessively repetitive patterns by focusing too much on restoring fine details, often neglecting the overall context. In contrast, models with refined pruned weights preserve the details of the pre-trained model while faithfully restoring its overall context. This result shows that, to achieve both diversity and fidelity in the compressed model, it is essential not only to preserve knowledge from the pre-trained model but also to refine the pruned weights for a proper balance with fine-tuning.

\subsection{Ablation Study on Other Pruning Method}

\begin{table}[t!]
\centering\resizebox{\linewidth}{!}
{
    \begin{tabular}{c|ccccc}
    \hline
    Method & FID $\downarrow$ & P $\uparrow$ & R $\uparrow$ & D $\uparrow$ & C $\uparrow$  \\ 
    \hline \hline

    CAGAN \cite{liu2021content} & 9.23 & 0.685 & 0.500 & 0.700 & 0.722 \\
    CAGAN (scaled) & \textbf{6.83} & \textbf{0.720} & \textbf{0.511} & \textbf{0.894} & \textbf{0.768} \\
    
    \hline
    \end{tabular}
}
\caption{Ablation study on the CAGAN with FFHQ. ``scaled'' means that SVS is applied to the pruned weights. }
\label{table:cagan-ablation}
\end{table}
\noindent There are only a few pruning methods for generative models, with CAGAN \cite{liu2021content}, DCP-GAN, and Diff-Prune considered key baselines. We extend our analysis to CAGAN to further validate the applicability of SVS beyond state-of-the-art methods. The pruned weights of CAGAN exhibit a similar singular value distribution to DCP-GAN, with existence of dominant singular vectors. As shown in \Tref{table:cagan-ablation}, applying SVS to CAGAN consistently leads to substantial improvements in all metrics, confirming that applicability of our method.

\subsection{Ablation Study on Different Scaling Functions}
\begin{table}[t!]
\centering\resizebox{\linewidth}{!}
{ 
    \begin{tabular}{c|c|ccccc}
    \hline
    Dataset & Scaling functions & FID $\downarrow$ & P $\uparrow$ & R $\uparrow$ & D $\uparrow$ & C $\uparrow$  \\ 
    \hline \hline

    \multirow{3}{*}{FFHQ} & $\log(x+1)$ & \textbf{6.18} & \textbf{0.732} & 0.513 & \underline{0.960} & \underline{0.789} \\
    & $\vert\log(x)\vert$ & \underline{6.32} & \underline{0.731} & \underline{0.519} & \textbf{0.975} & \textbf{0.796} \\
    & $\sqrt{x}$ & 6.60 & 0.719 & \textbf{0.532} & 0.874 & 0.778  \\
    \hline
    \multirow{3}{*}{Church} & $\log(x+1)$ & 5.30 & \textbf{0.723} & \underline{0.458} & \underline{0.933} & \textbf{0.809} \\
    & $\vert\log(x)\vert$ & 5.43 & \underline{0.720} & 0.453 & 0.916 & 0.799 \\
    & $\sqrt{x}$ & \textbf{5.18} & {0.719} & \textbf{0.464} & \textbf{0.934} & \underline{0.805}  \\
    \hline
    
    \end{tabular}

}
\caption{Ablation on the DCP-GAN (StyleGAN2 (small)) with different scaling functions.}
\label{table:scaling-function-ablation}
\end{table}

\noindent While we primarily use the square root function, we also test alternative scaling functions, such as $\log(x+1)$ and $|\log(x)|$, for reducing singular value gaps. \Tref{table:scaling-function-ablation} shows similar performance across all functions, highlighting the importance of addressing singular value gaps. The square root function was chosen for its simplicity, with results suggesting that broader exploration of scaling strategies could further enhance compression performance.

\subsection{Ablation Study on Different Scaling Methods}

\begin{table}[t!]
\centering\resizebox{0.9\linewidth}{!}
{
    \begin{tabular}{c|ccccc}
    \hline
    Method (StyleGAN2 (small), FFHQ) & FID $\downarrow$  \\ 
    \hline \hline
    (\textbf{a}) Normalizing singular values &  N/A \\ 
    (\textbf{b}) Spectral normalizing singular values &  N/A \\
    (\textbf{c}) Squaring singular values &  N/A \\ 
    (\textbf{d}) Spectral normalization to the generator & 8.00 \\ 
    (\textbf{e}) Iterative refining the generator's weights &  8.40 \\ 
    \textbf{\ours~(SVS)} &  \textbf{6.60} \\ 
    \hline
    \end{tabular}
}
\caption{Ablation on the different refinement methods. ``N/A'' indicates that the model diverges.}
\label{table:fid-stylegan2-ablation}
\end{table}

In \Tref{table:fid-stylegan2-ablation}, we explore additional approaches to refine singular values: (\textbf{a}) normalizing singular values (setting $\sigma_i = 1$), (\textbf{b}) normalizing with the spectral norm (dividing all singular values $\sigma_i$ by the largest singular value $\sigma_1$), (\textbf{c}) Squaring singular values, (\textbf{d}) applying spectral normalization to the synthesis network \cite{miyato2018spectral}, and (\textbf{e}) iterative refining the generator's weights with our method. For (\textbf{e}), we perform iterative refining only for the first half of the training due to performance degradation. 
From (\textbf{a}) and (\textbf{b}), we find that setting singular values to 1 (where the weights are orthogonal) or spectral normalizing singular values cause the training to diverge, similar to previous observations~\cite{saxe2014exact, wang2023trainability}. From (\textbf{c}), we find that widening singular value gaps leads to early training divergence, highlighting the importance of minimizing these gaps for effective fine-tuning. From (\textbf{d}), we see that applying spectral normalization to the generator’s weights makes it difficult for the generator to learn. Finally, from (\textbf{e}), we find that repeatedly refining the singular values of weights during training perturbs the learning process, depriving the generator of sufficient training time and leading to suboptimal results. From these results, we find that applying our method once at model initialization is not only efficient but also leads to more effective training. 

\section{Conclusion}

In this paper, we propose ``Singular Value Scaling" that refines pruned weights to achieve more effective and efficient generative model compression. Our method minimizes the disparities  between singular values of pruned weights, allowing the pruned model to explore a broader weight space. Our method is not only simple and effective but also versatile, applicable across different model types, GANs and Diffusion models. Extensive experiments demonstrate that our method provides a more effective initialization state, leading to performance improvements across various metrics for both StyleGAN and Diffusion model compression.

\bibliography{reference}

\begin{thebibliography}{44}
\providecommand{\natexlab}[1]{#1}
\providecommand{\url}[1]{\texttt{#1}}
\expandafter\ifx\csname urlstyle\endcsname\relax
  \providecommand{\doi}[1]{doi: #1}\else
  \providecommand{\doi}{doi: \begingroup \urlstyle{rm}\Url}\fi

\bibitem[Alex(2009)]{alex2009learning}
K.~Alex.
\newblock Learning multiple layers of features from tiny images.
\newblock \emph{https://www. cs. toronto. edu/kriz/learning-features-2009-TR. pdf}, 2009.

\bibitem[Baykal et~al.(2023)Baykal, Anees, Ceylan, Erdem, Erdem, and Yuret]{CLIPInverter}
A.~C. Baykal, A.~B. Anees, D.~Ceylan, E.~Erdem, A.~Erdem, and D.~Yuret.
\newblock Clip-guided stylegan inversion for text-driven real image editing.
\newblock \emph{ACM Trans. Graph.}, jul 2023.
\newblock ISSN 0730-0301.
\newblock \doi{10.1145/3610287}.
\newblock URL \url{https://doi.org/10.1145/3610287}.
\newblock Just Accepted.

\bibitem[Blattmann et~al.(2023)Blattmann, Rombach, Ling, Dockhorn, Kim, Fidler, and Kreis]{blattmann2023videoldm}
A.~Blattmann, R.~Rombach, H.~Ling, T.~Dockhorn, S.~W. Kim, S.~Fidler, and K.~Kreis.
\newblock Align your latents: High-resolution video synthesis with latent diffusion models.
\newblock In \emph{IEEE Conference on Computer Vision and Pattern Recognition ({CVPR})}, 2023.

\bibitem[Chan et~al.(2022)Chan, Lin, Chan, Nagano, Pan, Mello, Gallo, Guibas, Tremblay, Khamis, Karras, and Wetzstein]{Chan2022}
E.~R. Chan, C.~Z. Lin, M.~A. Chan, K.~Nagano, B.~Pan, S.~D. Mello, O.~Gallo, L.~Guibas, J.~Tremblay, S.~Khamis, T.~Karras, and G.~Wetzstein.
\newblock Efficient geometry-aware {3D} generative adversarial networks.
\newblock In \emph{CVPR}, 2022.

\bibitem[Chung et~al.(2024)Chung, Hyun, Shim, and Heo]{chung2024diversity}
J.~Chung, S.~Hyun, S.-H. Shim, and J.-P. Heo.
\newblock Diversity-aware channel pruning for stylegan compression.
\newblock \emph{arXiv preprint arXiv:2403.13548}, 2024.

\bibitem[Denton et~al.(2014)Denton, Zaremba, Bruna, LeCun, and Fergus]{denton2014exploiting}
E.~L. Denton, W.~Zaremba, J.~Bruna, Y.~LeCun, and R.~Fergus.
\newblock Exploiting linear structure within convolutional networks for efficient evaluation.
\newblock \emph{Advances in neural information processing systems}, 27, 2014.

\bibitem[Fang et~al.(2023)Fang, Ma, and Wang]{fang2023structural}
G.~Fang, X.~Ma, and X.~Wang.
\newblock Structural pruning for diffusion models.
\newblock In \emph{Advances in Neural Information Processing Systems}, 2023.

\bibitem[Girshick(2015)]{girshick2015fast}
R.~Girshick.
\newblock Fast r-cnn.
\newblock In \emph{Proceedings of the IEEE international conference on computer vision}, pages 1440--1448, 2015.

\bibitem[Glorot and Bengio(2010)]{glorot2010understanding}
X.~Glorot and Y.~Bengio.
\newblock Understanding the difficulty of training deep feedforward neural networks.
\newblock In \emph{Proceedings of the thirteenth international conference on artificial intelligence and statistics}, pages 249--256. JMLR Workshop and Conference Proceedings, 2010.

\bibitem[Goodfellow et~al.(2014)Goodfellow, Pouget-Abadie, Mirza, Xu, Warde-Farley, Ozair, Courville, and Bengio]{goodfellow2014generative}
I.~Goodfellow, J.~Pouget-Abadie, M.~Mirza, B.~Xu, D.~Warde-Farley, S.~Ozair, A.~Courville, and Y.~Bengio.
\newblock Generative adversarial nets.
\newblock \emph{Advances in neural information processing systems}, 27, 2014.

\bibitem[He et~al.(2015)He, Zhang, Ren, and Sun]{he2015delving}
K.~He, X.~Zhang, S.~Ren, and J.~Sun.
\newblock Delving deep into rectifiers: Surpassing human-level performance on imagenet classification.
\newblock In \emph{Proceedings of the IEEE international conference on computer vision}, pages 1026--1034, 2015.

\bibitem[Heusel et~al.(2017)Heusel, Ramsauer, Unterthiner, Nessler, and Hochreiter]{heusel2017gans}
M.~Heusel, H.~Ramsauer, T.~Unterthiner, B.~Nessler, and S.~Hochreiter.
\newblock Gans trained by a two time-scale update rule converge to a local nash equilibrium.
\newblock \emph{Advances in neural information processing systems}, 30, 2017.

\bibitem[Ho et~al.(2020)Ho, Jain, and Abbeel]{ho2020denoising}
J.~Ho, A.~Jain, and P.~Abbeel.
\newblock Denoising diffusion probabilistic models.
\newblock \emph{Advances in neural information processing systems}, 33:\penalty0 6840--6851, 2020.

\bibitem[Ho et~al.(2022)Ho, Salimans, Gritsenko, Chan, Norouzi, and Fleet]{ho2022video}
J.~Ho, T.~Salimans, A.~Gritsenko, W.~Chan, M.~Norouzi, and D.~J. Fleet.
\newblock Video diffusion models.
\newblock \emph{Advances in Neural Information Processing Systems}, 35:\penalty0 8633--8646, 2022.

\bibitem[Karnewar et~al.(2023)Karnewar, Vedaldi, Novotny, and Mitra]{Karnewar_2023_CVPR}
A.~Karnewar, A.~Vedaldi, D.~Novotny, and N.~J. Mitra.
\newblock Holodiffusion: Training a 3d diffusion model using 2d images.
\newblock In \emph{Proceedings of the IEEE/CVF Conference on Computer Vision and Pattern Recognition (CVPR)}, pages 18423--18433, June 2023.

\bibitem[Karras et~al.(2019)Karras, Laine, and Aila]{stylegan1}
T.~Karras, S.~Laine, and T.~Aila.
\newblock A style-based generator architecture for generative adversarial networks.
\newblock In \emph{2019 IEEE/CVF Conference on Computer Vision and Pattern Recognition (CVPR)}, pages 4396--4405, 2019.
\newblock \doi{10.1109/CVPR.2019.00453}.

\bibitem[Karras et~al.(2020)Karras, Laine, Aittala, Hellsten, Lehtinen, and Aila]{Karras2019stylegan2}
T.~Karras, S.~Laine, M.~Aittala, J.~Hellsten, J.~Lehtinen, and T.~Aila.
\newblock Analyzing and improving the image quality of {StyleGAN}.
\newblock In \emph{Proc. CVPR}, 2020.

\bibitem[Karras et~al.(2021)Karras, Aittala, Laine, H{\"a}rk{\"o}nen, Hellsten, Lehtinen, and Aila]{karras2021alias}
T.~Karras, M.~Aittala, S.~Laine, E.~H{\"a}rk{\"o}nen, J.~Hellsten, J.~Lehtinen, and T.~Aila.
\newblock Alias-free generative adversarial networks.
\newblock \emph{Advances in neural information processing systems}, 34:\penalty0 852--863, 2021.

\bibitem[Kawar et~al.(2023)Kawar, Zada, Lang, Tov, Chang, Dekel, Mosseri, and Irani]{kawar2023imagic}
B.~Kawar, S.~Zada, O.~Lang, O.~Tov, H.~Chang, T.~Dekel, I.~Mosseri, and M.~Irani.
\newblock Imagic: Text-based real image editing with diffusion models.
\newblock In \emph{Conference on Computer Vision and Pattern Recognition 2023}, 2023.

\bibitem[Kingma and Ba(2015)]{2015-kingma}
D.~P. Kingma and J.~Ba.
\newblock Adam: A method for stochastic optimization.
\newblock In Y.~Bengio and Y.~LeCun, editors, \emph{ICLR (Poster)}, 2015.
\newblock URL \url{http://dblp.uni-trier.de/db/conf/iclr/iclr2015.html#KingmaB14}.

\bibitem[Kynk{\"a}{\"a}nniemi et~al.(2019)Kynk{\"a}{\"a}nniemi, Karras, Laine, Lehtinen, and Aila]{kynkaanniemi2019improved}
T.~Kynk{\"a}{\"a}nniemi, T.~Karras, S.~Laine, J.~Lehtinen, and T.~Aila.
\newblock Improved precision and recall metric for assessing generative models.
\newblock \emph{Advances in Neural Information Processing Systems}, 32, 2019.

\bibitem[Liu et~al.(2021)Liu, Shu, Li, Lin, Perazzi, and Kung]{liu2021content}
Y.~Liu, Z.~Shu, Y.~Li, Z.~Lin, F.~Perazzi, and S.-Y. Kung.
\newblock {Content-Aware GAN Compression}.
\newblock In \emph{CVPR}, 2021.

\bibitem[Liu et~al.(2015)Liu, Luo, Wang, and Tang]{liu2015faceattributes}
Z.~Liu, P.~Luo, X.~Wang, and X.~Tang.
\newblock Deep learning face attributes in the wild.
\newblock In \emph{Proceedings of International Conference on Computer Vision (ICCV)}, December 2015.

\bibitem[Lu et~al.(2022)Lu, Zhou, Bao, Chen, Li, and Zhu]{lu2022dpmsolver}
C.~Lu, Y.~Zhou, F.~Bao, J.~Chen, C.~Li, and J.~Zhu.
\newblock {DPM}-solver: A fast {ODE} solver for diffusion probabilistic model sampling in around 10 steps.
\newblock In A.~H. Oh, A.~Agarwal, D.~Belgrave, and K.~Cho, editors, \emph{Advances in Neural Information Processing Systems}, 2022.
\newblock URL \url{https://openreview.net/forum?id=2uAaGwlP_V}.

\bibitem[Mescheder et~al.(2018)Mescheder, Geiger, and Nowozin]{mescheder2018training}
L.~Mescheder, A.~Geiger, and S.~Nowozin.
\newblock Which training methods for gans do actually converge?
\newblock In \emph{International conference on machine learning}, pages 3481--3490. PMLR, 2018.

\bibitem[Miyato et~al.(2018)Miyato, Kataoka, Koyama, and Yoshida]{miyato2018spectral}
T.~Miyato, T.~Kataoka, M.~Koyama, and Y.~Yoshida.
\newblock Spectral normalization for generative adversarial networks.
\newblock In \emph{International Conference on Learning Representations}, 2018.
\newblock URL \url{https://openreview.net/forum?id=B1QRgziT-}.

\bibitem[Naeem et~al.(2020)Naeem, Oh, Uh, Choi, and Yoo]{naeem2020reliable}
M.~F. Naeem, S.~J. Oh, Y.~Uh, Y.~Choi, and J.~Yoo.
\newblock Reliable fidelity and diversity metrics for generative models.
\newblock In \emph{International Conference on Machine Learning}, pages 7176--7185. PMLR, 2020.

\bibitem[Pehlivan et~al.(2023)Pehlivan, Dalva, and Dundar]{Pehlivan_2023_CVPR}
H.~Pehlivan, Y.~Dalva, and A.~Dundar.
\newblock Styleres: Transforming the residuals for real image editing with stylegan.
\newblock In \emph{Proceedings of the IEEE/CVF Conference on Computer Vision and Pattern Recognition (CVPR)}, pages 1828--1837, June 2023.

\bibitem[Petersen(2012)]{petersen2012linear}
P.~Petersen.
\newblock \emph{Linear algebra}.
\newblock Springer, 2012.

\bibitem[Saxe et~al.(2014)Saxe, McClelland, and Ganguli]{saxe2014exact}
A.~Saxe, J.~McClelland, and S.~Ganguli.
\newblock Exact solutions to the nonlinear dynamics of learning in deep linear neural networks.
\newblock In \emph{Proceedings of the International Conference on Learning Represenatations 2014}. International Conference on Learning Represenatations 2014, 2014.

\bibitem[Skorokhodov et~al.(2022)Skorokhodov, Tulyakov, and Elhoseiny]{stylegan-v}
I.~Skorokhodov, S.~Tulyakov, and M.~Elhoseiny.
\newblock Stylegan-v: A continuous video generator with the price, image quality and perks of stylegan2.
\newblock In \emph{Proceedings of the IEEE/CVF Conference on Computer Vision and Pattern Recognition}, pages 3626--3636, 2022.

\bibitem[Song et~al.(2021)Song, Meng, and Ermon]{song2021denoising}
J.~Song, C.~Meng, and S.~Ermon.
\newblock Denoising diffusion implicit models.
\newblock In \emph{International Conference on Learning Representations}, 2021.
\newblock URL \url{https://openreview.net/forum?id=St1giarCHLP}.

\bibitem[Wang and Fu(2023)]{wang2023trainability}
H.~Wang and Y.~Fu.
\newblock Trainability preserving neural pruning.
\newblock In \emph{The Eleventh International Conference on Learning Representations}, 2023.
\newblock URL \url{https://openreview.net/forum?id=AZFvpnnewr}.

\bibitem[Wang et~al.(2020)Wang, Gui, Yang, Liu, and Wang]{wang2020ganslimming}
H.~Wang, S.~Gui, H.~Yang, J.~Liu, and Z.~Wang.
\newblock Gan slimming: All-in-one gan compression by a unified optimization framework.
\newblock In \emph{European Conference on Computer Vision}, 2020.

\bibitem[Wang et~al.(2022)Wang, Yang, Xu, Shen, Li, and Zhou]{Wang_2022_CVPR}
J.~Wang, C.~Yang, Y.~Xu, Y.~Shen, H.~Li, and B.~Zhou.
\newblock Improving gan equilibrium by raising spatial awareness.
\newblock In \emph{Proceedings of the IEEE/CVF Conference on Computer Vision and Pattern Recognition (CVPR)}, pages 11285--11293, June 2022.

\bibitem[Wang et~al.(2004)Wang, Bovik, Sheikh, and Simoncelli]{wang2004image}
Z.~Wang, A.~C. Bovik, H.~R. Sheikh, and E.~P. Simoncelli.
\newblock Image quality assessment: from error visibility to structural similarity.
\newblock \emph{IEEE transactions on image processing}, 13\penalty0 (4):\penalty0 600--612, 2004.

\bibitem[Xu et~al.(2022)Xu, Hou, Liu, and Loy]{xu2022mind}
G.~Xu, Y.~Hou, Z.~Liu, and C.~C. Loy.
\newblock Mind the gap in distilling stylegans.
\newblock In \emph{European Conference on Computer Vision}, pages 423--439. Springer, 2022.

\bibitem[Yeo et~al.(2024)Yeo, Jang, and Yoo]{yeo2024nickeldiminggandualmethod}
S.~Yeo, Y.~Jang, and J.~Yoo.
\newblock Nickel and diming your gan: A dual-method approach to enhancing gan efficiency via knowledge distillation, 2024.
\newblock URL \url{https://arxiv.org/abs/2405.11614}.

\bibitem[Yoo et~al.(2019)Yoo, Uh, Chun, Kang, and Ha]{yoo2019photorealistic}
J.~Yoo, Y.~Uh, S.~Chun, B.~Kang, and J.-W. Ha.
\newblock Photorealistic style transfer via wavelet transforms.
\newblock In \emph{Proceedings of the IEEE/CVF international conference on computer vision}, pages 9036--9045, 2019.

\bibitem[Yu et~al.(2015)Yu, Seff, Zhang, Song, Funkhouser, and Xiao]{yu2015lsun}
F.~Yu, A.~Seff, Y.~Zhang, S.~Song, T.~Funkhouser, and J.~Xiao.
\newblock Lsun: Construction of a large-scale image dataset using deep learning with humans in the loop.
\newblock \emph{arXiv preprint arXiv:1506.03365}, 2015.

\bibitem[Zhang et~al.(2018)Zhang, Isola, Efros, Shechtman, and Wang]{zhang2018unreasonable}
R.~Zhang, P.~Isola, A.~A. Efros, E.~Shechtman, and O.~Wang.
\newblock The unreasonable effectiveness of deep features as a perceptual metric.
\newblock In \emph{Proceedings of the IEEE conference on computer vision and pattern recognition}, pages 586--595, 2018.

\bibitem[Zhang et~al.(2015)Zhang, Zou, Ming, He, and Sun]{zhang2015efficient}
X.~Zhang, J.~Zou, X.~Ming, K.~He, and J.~Sun.
\newblock Efficient and accurate approximations of nonlinear convolutional networks.
\newblock In \emph{Proceedings of the IEEE Conference on Computer Vision and pattern Recognition}, pages 1984--1992, 2015.

\bibitem[Zhang et~al.(2023)Zhang, Han, Ghosh, Metaxas, and Ren]{zhang2023sine}
Z.~Zhang, L.~Han, A.~Ghosh, D.~N. Metaxas, and J.~Ren.
\newblock Sine: Single image editing with text-to-image diffusion models.
\newblock In \emph{Proceedings of the IEEE/CVF Conference on Computer Vision and Pattern Recognition}, pages 6027--6037, 2023.

\bibitem[Zheng et~al.(2023)Zheng, Nie, Vahdat, Azizzadenesheli, and Anandkumar]{zheng2023fast}
H.~Zheng, W.~Nie, A.~Vahdat, K.~Azizzadenesheli, and A.~Anandkumar.
\newblock Fast sampling of diffusion models via operator learning.
\newblock In \emph{International conference on machine learning}, pages 42390--42402. PMLR, 2023.

\end{thebibliography}

\onecolumn
\begin{center}
{\LARGE\bfseries \ours: Efficient Generative Model Compression via Pruned Weights Refinement\\- Supplementary Materials -}
\end{center}

\section{Implementation Details}
\paragraph{StyleGAN2} We first prune the pre-trained model using the DCP-GAN \cite{chung2024diversity} method with 70\% channel sparsity. We then refine the pruned weights with our method.  Basically, we follow the training scheme of the DCP-GAN, which shares same training scheme with StyleKD \cite{xu2022mind}. In detail, we employ non-saturating loss, $\mathcal L_{GAN}$ \cite{goodfellow2014generative} with R1 regularization loss \cite{mescheder2018training, stylegan1}, optimizing the model with Adam optimizer \cite{2015-kingma} with $\beta_1=0.0$ and $\beta_2=0.99$. The the batch size and learning rate are set to 32 and 2e-3 for the StyleGAN2 (base) architecture, while set to 64 and 2.5e-3 for the StyeGAN2 (small) following official StyleGAN2 implementation. For all experiments, we consistently employ 4 GPUs to train the models. For the knowledge distillation, we adopt $\mathcal{L}_{rgb}$ (pixel loss), $\mathcal{L}_{lpips}$ (LPIPS loss \cite{zhang2018unreasonable}), and $\mathcal{L}_{LD}$ (latent-direction-based distillation loss \cite{xu2022mind}), with the hyperparameters, $\lambda_{GAN}=1,\ \lambda_{rgb}=3,\ \lambda_{lpips}=3,\ \lambda_{LD}=30 $ respectively. To match the training iterations with previous studies, we train the models until the discriminator sees 15000K images for StyleGAN2 (base), which corresponds to approximately 470,000 training iterations. For StyleGAN2 (small), we train the models until the discriminator sees 25000K images, which corresponds to approximately 390,000 training iterations.

\paragraph{StyleGAN3} For StyleGAN3 compression, we implement our method based on the StyleGAN3 compression implementation from DCP-GAN \cite{chung2024diversity}. Similar to StyleGAN2 compression scheme, we first prune the pre-trained model using DCP-GAN method with 70\% channel sparsity and refine the pruned weights with our method. The same training losses and hyperparameters used for StyleGAN2 compression are employed here as well. The model is optimized with Adam optimizer with  $\beta_1=0.0$ and $\beta_2=0.99$. The batch size is set to 16 following DCP-GAN, and the learning rate is set to 2.5e-3 for the generator and set to 2e-3 for the discriminator. We change the layer outputs used in $\mathcal L_{LD}$ from \{layer\_1,\ layer\_2\} (used in the DCP-GAN implementation) to \{layer\_2, layer\_4, layer\_6, layer\_9\} to match the number of layers used in $\mathcal L_{LD}$ with StyleGAN2 compression.

\paragraph{DDPM} For Diffusion model compression, we first prune the pre-trained model using Diff-Prune \cite{fang2023structural} method with varying pruning ratios and then refine the pruned weights with our method. And then, the pruned models are trained with noise prediction loss, proposed in DDPM \cite{ho2020denoising}. The batch size and learning rate are set to 128, 2e-4 for CIFAR10, 96, 2e-4 for CelebA and 64, optimizing the models with Adam optimizer \cite{2015-kingma} with $\beta_1=0.9$ and $\beta_2=0.999$, following the official implementation of Diff-Prune.

\section{Reproducibility and Benchmarking}
\paragraph{Reproducibility} Naively adopting the implementation of StyleKD on the official StyleGAN2 implementation will fail to reproduce the results. This problem is caused by the use of mixed precision operations in the official StyleGAN2 implementations, where the intermediate layer output type is float16. If we directly use the intermediate layer output features for $\mathcal{L}_{LD}$ (latent-direction-based distillation loss \cite{xu2022mind}), the back-propagated gradients become inaccurate, leading to divergence in training. Therefore, we need to convert these intermediate features to float32 before using them in the loss function to ensure stable model training.

\paragraph{Benchmarking} In this paper, we conduct a comprehensive benchmarking of existing StyleGAN2 compression methods using the official StyleGAN2 implementation\footnote{https://github.com/NVlabs/stylegan2-ada-pytorch}. In concurrent work, Nickel \& DiME \cite{yeo2024nickeldiminggandualmethod} have similarly reproduced previous StyleGAN2 compression methods using the official implementation. However, they failed to reproduce the state-of-the-art compression methods, such as StyleKD and DCP-GAN. With our implementation, the performance of the DCP-GAN \cite{chung2024diversity}, can be faithfully reproduced (FID=6.35 (reported) $\rightarrow$ FID=6.51 (reproduced) in the FFHQ dataset). One notable thing is that in our implementation, StyleKD \cite{xu2022mind} demonstrates significantly improved results (FID=7.25 (reported) $\rightarrow$ FID=6.70 (reproduced) in the FFHQ dataset.). We speculate that this performance improvement is due to the optimized GAN training in the official implementation. We hope that our implementation and benchmarks will provide a useful starting point for future research.

\section{Further Analysis}

\begin{table}[h!]
\centering
{
    \begin{tabular}{cccccc}
    \hline
    \multicolumn{6}{c}{\textbf{StyleGAN2 (small) FFHQ-256}} \\ \hline
    Method & FID $\downarrow$ & P $\uparrow$ & R $\uparrow$ & D $\uparrow$ & C $\uparrow$ \\ \hline \hline
    \makecell{StyleKD\\(He init)}  & \textbf{5.84} & \textbf{0.741} & 0.526 & \textbf{0.985} & \textbf{0.800} \\ 
    \makecell{DCP-GAN\\(scaled)} & 6.60 & 0.719 & \textbf{0.532} & 0.874 & 0.778 \\ \hline \hline
    \end{tabular}
    \begin{tabular}{cccccc}
    \hline
    \multicolumn{6}{c}{\textbf{StyleGAN2 (small) LSUN Church-256}} \\ \hline
    Method & FID $\downarrow$ & P $\uparrow$ & R $\uparrow$ & D $\uparrow$ & C $\uparrow$ \\ \hline \hline
    \makecell{StyleKD\\(He init)} & 5.45 & \textbf{0.725} & 0.448 & 0.924 & 0.799 \\ 
    \makecell{DCP-GAN\\(scaled)} & \textbf{5.18} & 0.719 & \textbf{0.464} & \textbf{0.934} & \textbf{0.805} \\ \hline \hline
    \end{tabular}
}

\caption{Comparison between pruned weights initialization and random weight initialization on StyleGAN2 (small). The channel sparsity is set to 70\%. ``He init'' denotes He initialization is appled to the randomly initialized weights. ``scaled'' means that our \textit{\ours} is applied to the pruned weights.}

\label{table:he-init}
\end{table}

\paragraph{Effect of Weight initialization technique}
Unlike other CNN-based models where weights are typically initialized with He initialization \cite{he2015delving} (e.g., torch.nn.init.kaiming\_uniform in torch.nn.Conv2d), the weights in StyleGAN families \cite{Karras2019stylegan2, karras2021alias} are initialized using a unit normal distribution (e.g., torch.nn.randn). We observe that applying He initialization to randomly initialized weights improves further improves fine-tuning process, leading to better performance (see \Tref{table:he-init}). This even outperforms pruned weights initialization in the FFHQ dataset. However, in the LSUN Church dataset, pruned weights initialization still outperforms random weight initialization. Similarly, in the case of Diffusion model compression \cite{fang2023structural}, random weight initialization yields significantly worse results compared to pruned weights initialization. Therefore, we consider this a special case limited to StyleGAN2 compression in the FFHQ dataset.

\paragraph{Ablation study on extreme compression ratio cases}

\begin{table*}[ht!]
\centering
{
    \begin{tabular}{c|c|c|cc|ccccc}
    \hline
    Dataset & Method & Arch & Params  $\downarrow$ & FLOPs $\downarrow$ & FID $\downarrow$ & P $\uparrow$ & R $\uparrow$ & D $\uparrow$ & C $\uparrow$ \\ \hline \hline
    
    \multirow{4}{*}{FFHQ}
        
    & Teacher & \multirow{4}{*}{\text{small}} & 24.7M & 14.9B & 4.02 & 0.769 & 0.555 & 1.095 & 0.854 \\ 
    & StyleKD \cite{xu2022mind} & & \multirow{3}{*}{2.6M} & \multirow{3}{*}{0.16B} & \underline{19.53} & \textbf{0.585} & \underline{0.274} & \underline{0.496} & \underline{0.508} \\
    & DCP-GAN \cite{chung2024diversity} &  &  &  & 21.14 & 0.546 & 0.272 & 0.429 & 0.462 \\ 
    & \textbf{DCP-GAN (scaled)} &  & &  & \textbf{17.60} & \underline{0.578} & \textbf{0.342} & \textbf{0.540} & \textbf{0.550} \\ \hline    
    \end{tabular}
}

\caption{Quantitative results on StyleGAN2 compression with extreme compression ratio. StyleGAN2 (small) is compressed with the different compression methods, with the channel sparsity set to 90\%. ``scaled'' means that our \textit{\ours} is applied to the pruned weights.}
\label{table:fid-stylegan2-extreme}
\end{table*}

In \Tref{table:fid-stylegan2-extreme}, we present the performance of previous methods and the proposed method at higher compression rates for StyleGAN2. Specifically, we increased the channel sparsity from the original 70\% to 90\%, resulting in a compression rate of 98.92\%. As can be seen in \Tref{table:fid-stylegan2-extreme}, our method demonstrates significantly better performance compared to previous methods, proving the superiority of our method. Similarly, in the case of DDPM compression, when we increased the channel sparsity to 90\%, we observe that the FID of the compressed model fails to converge.

\section{Additional Results}

\begin{figure*}[p]
    \centering
    
    \begin{minipage}{\linewidth}
        \centering
        \includegraphics[width=0.9\linewidth]{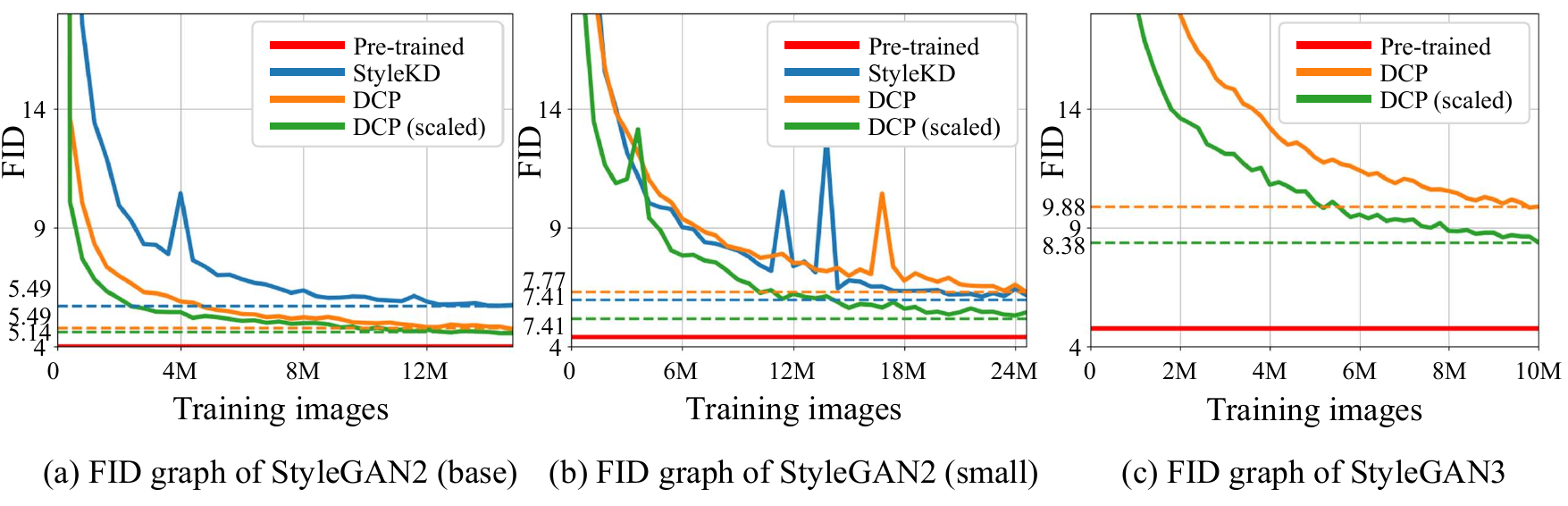}
        \caption{(a), (b) FID convergence graph in different StyleGAN2 Architectures compressed by different methods \cite{xu2022mind, chung2024diversity} in LSUN Church dataset, (c) FID convergence graph in StyleGAN3 Architecture compressed by different methods in FFHQ dataset. Solid line represents FID with respect to the training images, while dashed line represents the best FID of the compressed model by the corresponding method. ``scaled'' means that our \textit{\ours} is applied to the pruned weights. 
        } 
        \label{figure:suppl-stylegan-fid-convergence}
    \end{minipage}
    
    \vspace{0.5cm} 
    
    \begin{minipage}{\linewidth}
        \centering
        \includegraphics[width=0.9\linewidth]{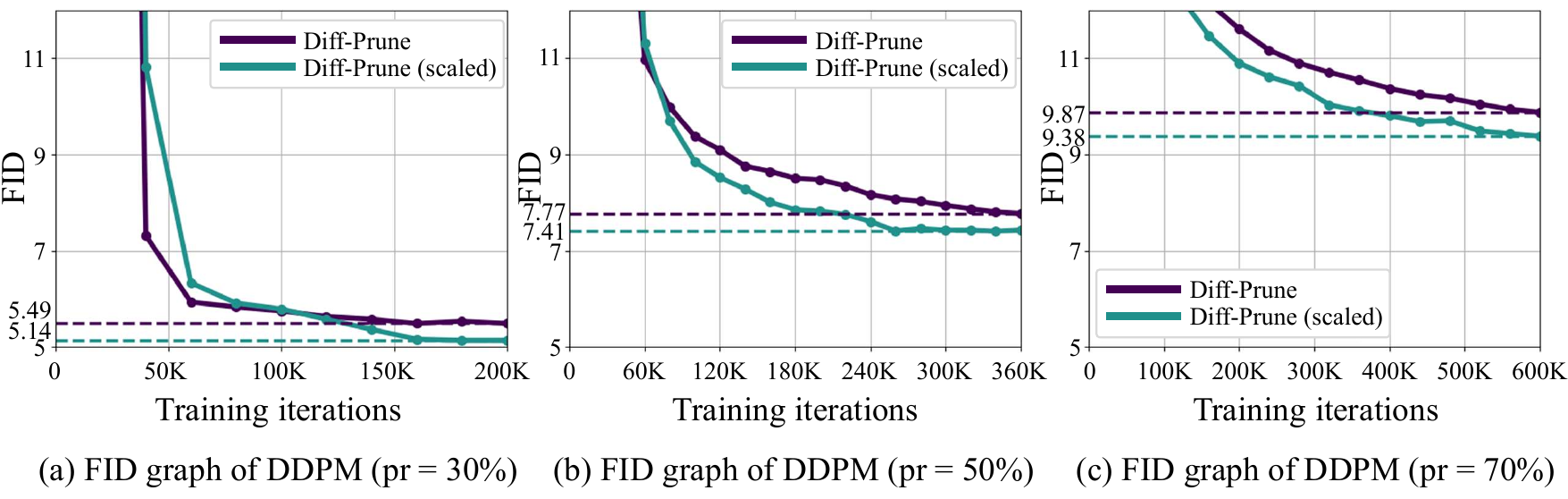} 
        \caption{FID convergence graph in DDPM compressed by different methods \cite{fang2023structural} in CIFAR10 dataset. Solid line represents FID with respect to the training images, while dashed line represents the best FID of the compressed model by the corresponding method. ``pr'' denotes the channel sparsity (pruning ratio). ``scaled'' means that our \textit{\ours} is applied to the pruned weights. 
        } 
        \label{figure:suppl-ddpm-cifar-fid-convergence}
    \end{minipage}
    
    \vspace{0.5cm} 
    
    \begin{minipage}{\linewidth}
        \centering
        \includegraphics[width=0.9\linewidth]{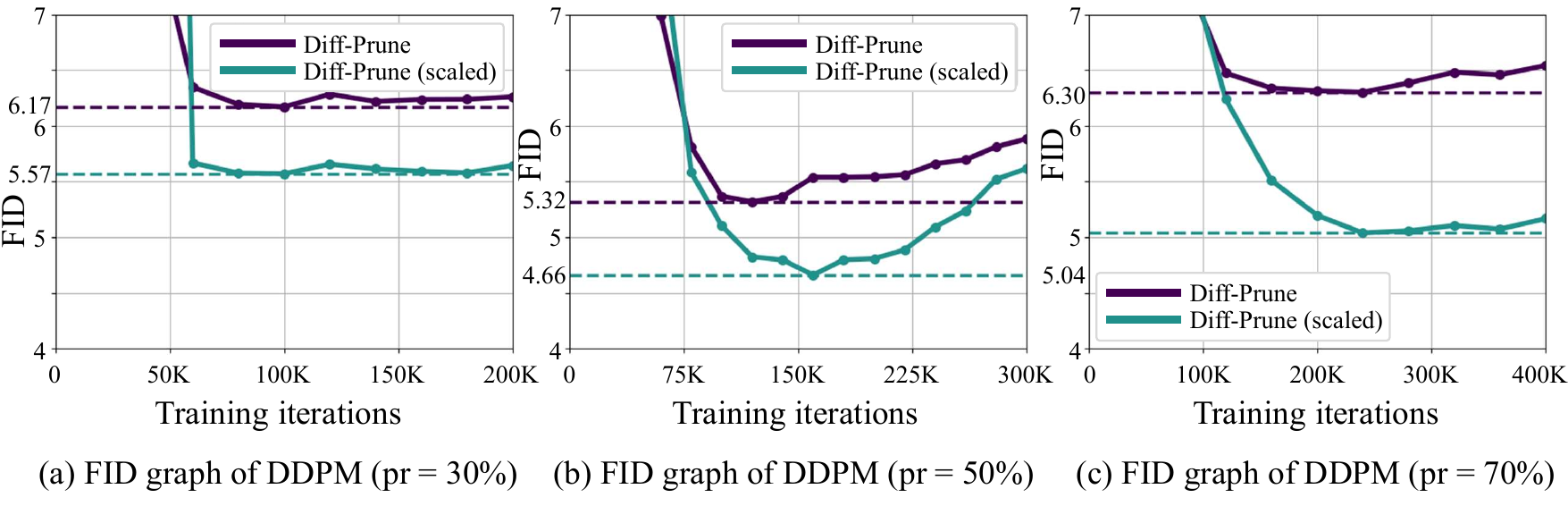} 
        \caption{FID convergence graph in DDPM compressed by different methods \cite{fang2023structural} in 
        CelebA-HQ dataset. Solid line represents FID with respect to the training images, while dashed line represents the best FID of the compressed model by the corresponding method. ``pr'' denotes the channel sparsity (pruning ratio). ``scaled'' means that our \textit{\ours} is applied to the pruned weights. 
        } 
        \label{figure:suppl-ddpm-celeba-fid-convergence}
    \end{minipage}
    \label{fig:fid-convergence-overall}
\end{figure*}

\paragraph{Additional FID convergence graph}
We provide more FID convergence graph in \Fref{figure:suppl-stylegan-fid-convergence} (StyleGAN2 in LSUN Church, StyleGAN3 in FFHQ), \Fref{figure:suppl-ddpm-cifar-fid-convergence} (DDPM in CIFAR10) and \Fref{figure:suppl-ddpm-celeba-fid-convergence} (DDPM in CelebA-HQ).

\begin{figure*}[p]
\centering
\includegraphics[width=\linewidth]{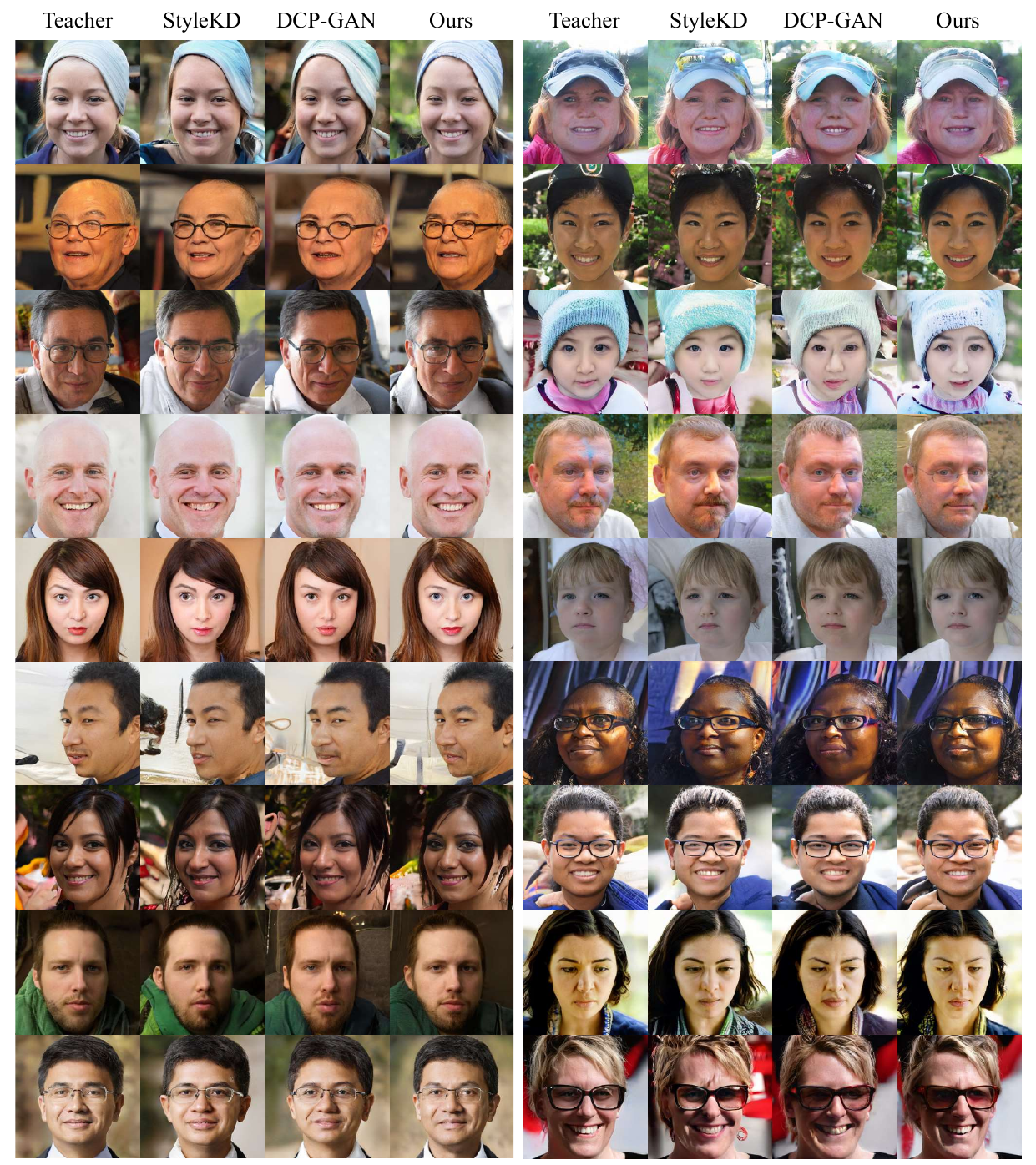} 
\caption{Qualitative results in StyleGAN2 (base) in FFHQ-256 dataset. All samples are uncurated. Each half row of samples is generated from same noise vector and the truncation trick parameter $\psi=1$ with StyleGAN2 (base) compressed using different compression methods with channel sparsity 70\%. ``Ours'' denotes the compressed model with DCP-GAN refined by \textit{\ours}.
} 
\label{figure:suppl-stylegan2-base-ffhq}
\end{figure*}

\begin{figure*}[p]
\centering
\includegraphics[width=\linewidth]{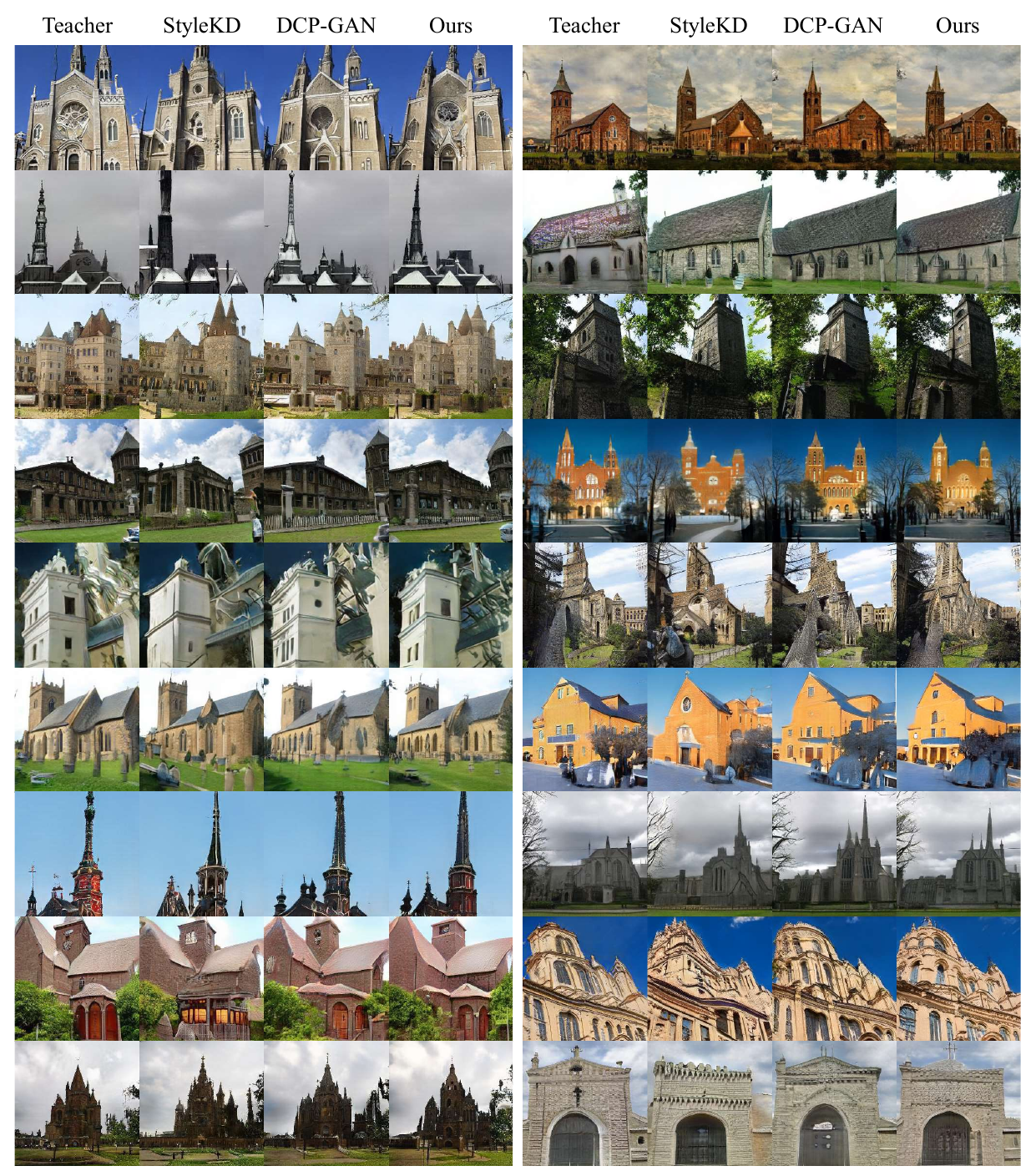} 
\caption{Qualitative results in StyleGAN2 (base) in LSUN Church-256 dataset. All samples are uncurated. Each half row of samples is generated from same noise vector and the truncation trick parameter $\psi=1$ with StyleGAN2 (base) compressed using different compression methods with channel sparsity 70\%. ``Ours'' denotes the compressed model with DCP-GAN refined by \textit{\ours}.} 
\label{figure:suppl-stylegan2-base-church}
\end{figure*}

\begin{figure*}[p]
\centering
\includegraphics[width=\linewidth]{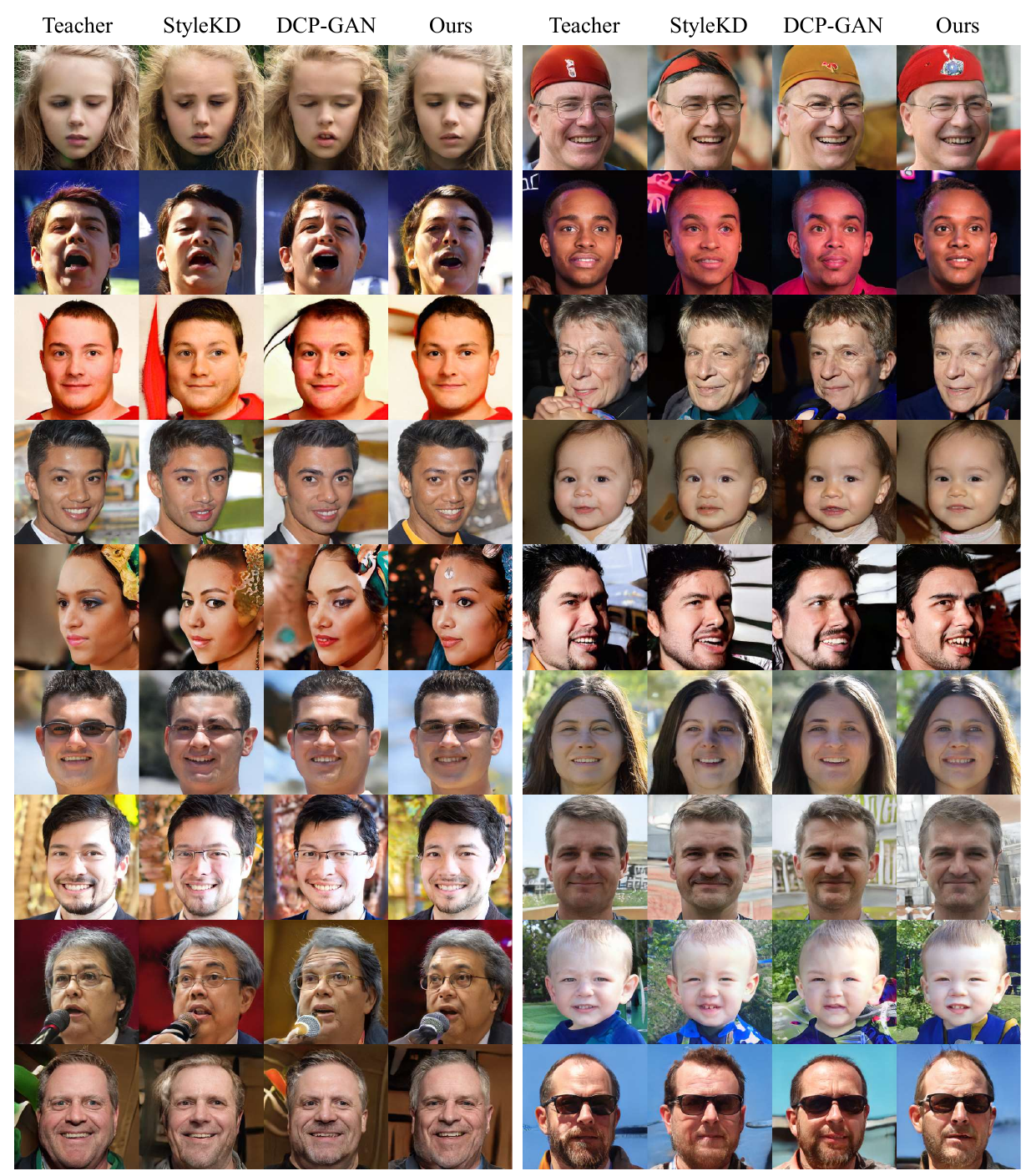}
\caption{Qualitative results in StyleGAN2 (small) in FFHQ-256 dataset. All samples are uncurated. Each half row of samples is generated from same noise vector and the truncation trick parameter $\psi=1$ with StyleGAN2 (base) compressed using different compression methods with channel sparsity 70\%. ``Ours'' denotes the compressed model with DCP-GAN refined by \textit{\ours}.} 
\label{figure:suppl-stylegan2-small-ffhq}
\end{figure*}

\begin{figure*}[p]
\centering
\includegraphics[width=\linewidth]{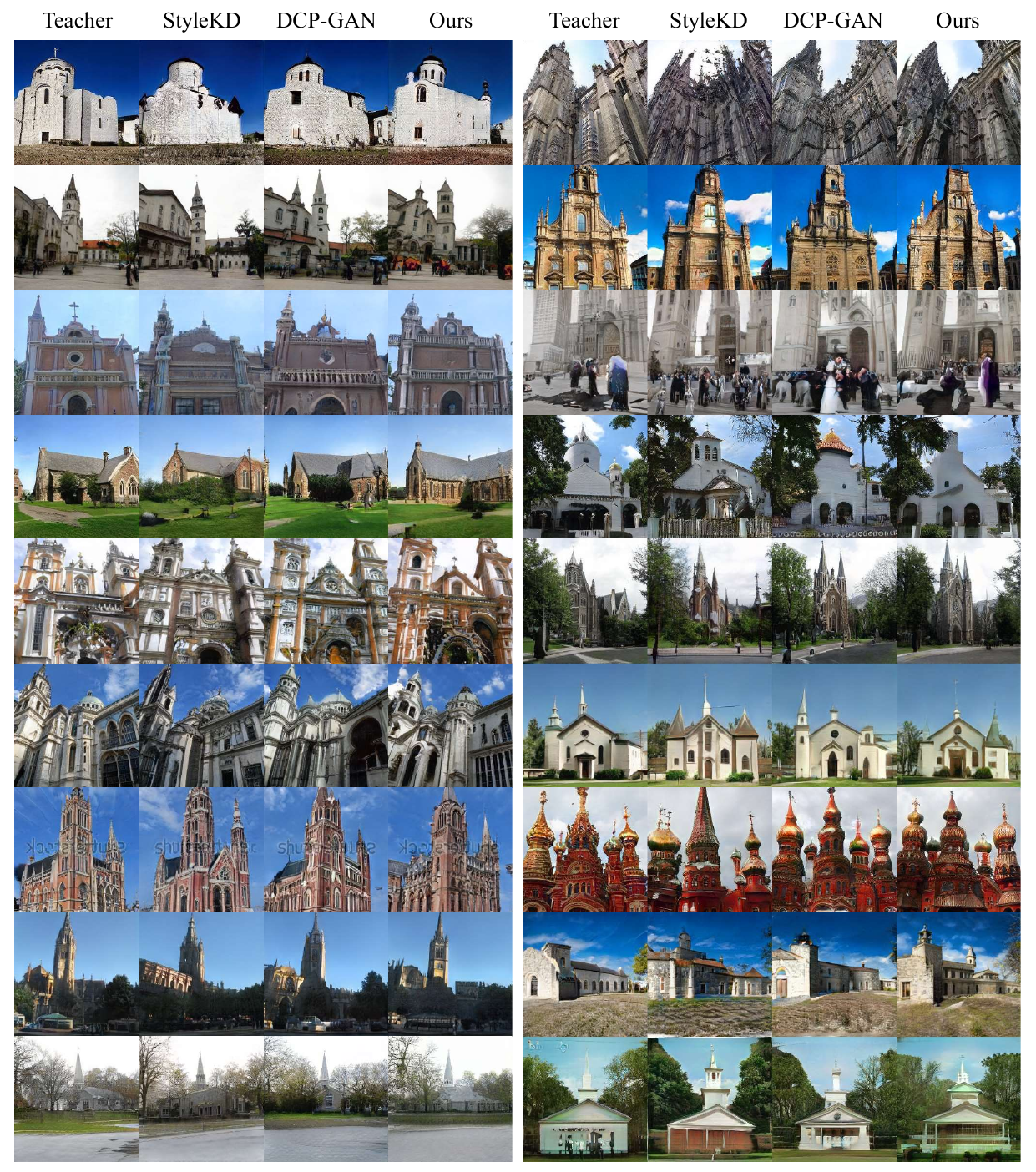} 
\caption{Qualitative results in StyleGAN2 (small) in LSUN Church-256 dataset. All samples are uncurated. Each half row of samples is generated from same noise vector and the truncation trick parameter $\psi=1$ with StyleGAN2 (base) compressed using different compression methods with channel sparsity 70\%. ``Ours'' denotes the compressed model with DCP-GAN refined by \textit{\ours}.
} 
\label{figure:suppl-stylegan2-small-church}
\end{figure*}

\begin{figure*}[p]
\centering
\includegraphics[width=\linewidth]{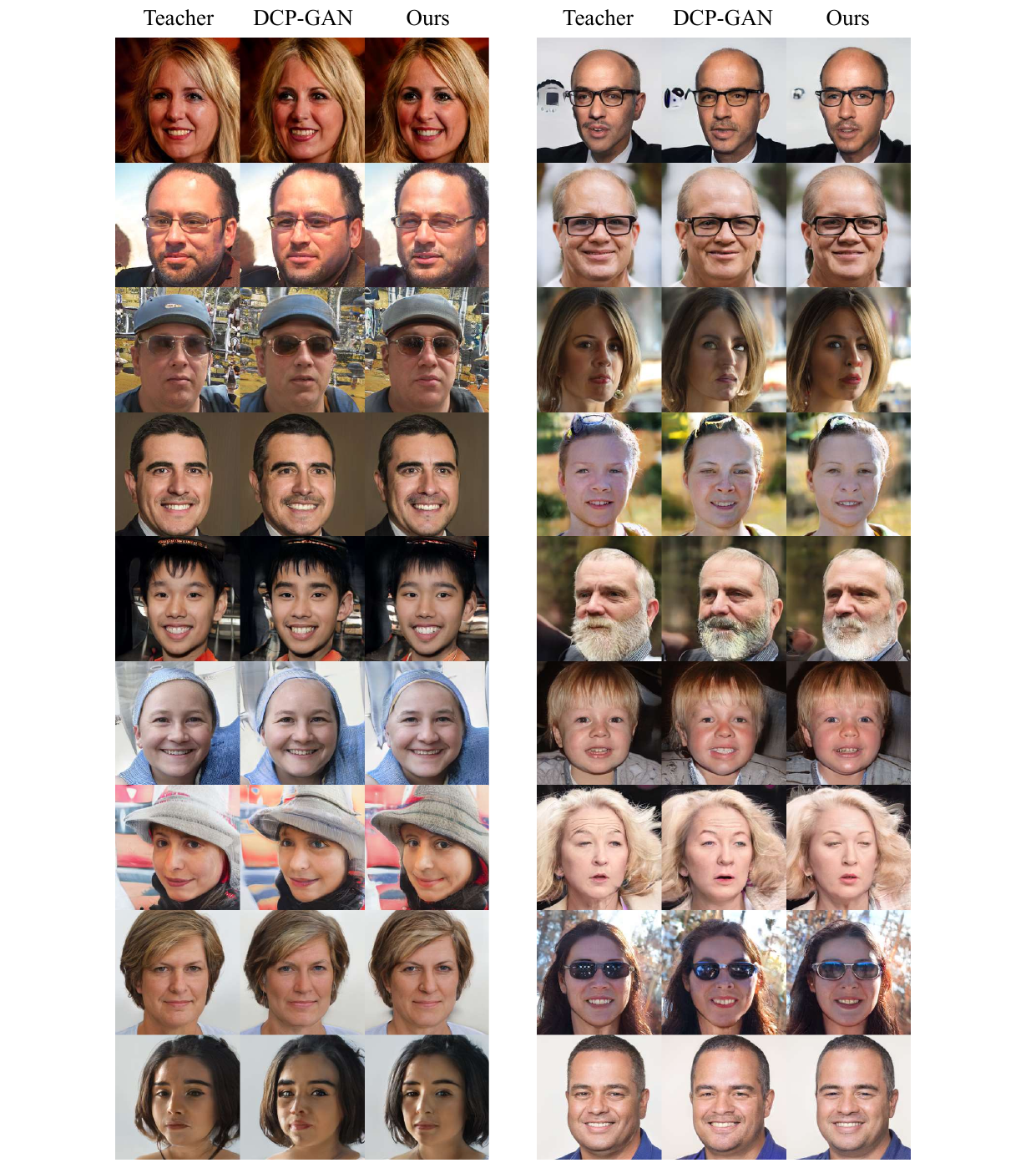} 
\caption{Qualitative results in StyleGAN3-T in FFHQ-256 dataset. All samples are uncurated. Each half row of samples is generated from same noise vector and the truncation trick parameter $\psi=1$ with StyleGAN2 (base) compressed using different compression methods with channel sparsity 70\%. ``Ours'' denotes the compressed model with DCP-GAN refined by \textit{\ours}.
} 
\label{figure:suppl-stylegan3-ffhq}
\end{figure*}

\paragraph{Additional qualitative results on StyleGAN compression}
We provide more qualitative comparison between the proposed method and previous methods \cite{xu2022mind, chung2024diversity}. Particularly, we provide the generated samples from the same for each methods in different StyleGAN2 architectures (StyleGAN (base), StyleGAN2 (small)) in FFHQ-256 and LSUN Church-256 datasets and StyleGAN3 in FFHQ-256. The generated samples can be found in \Fref{figure:suppl-stylegan2-base-ffhq}, \ref{figure:suppl-stylegan2-base-church}, \ref{figure:suppl-stylegan2-small-ffhq}, \ref{figure:suppl-stylegan2-small-church}, \ref{figure:suppl-stylegan3-ffhq}.

\end{document}